\documentclass{article}

\pdfoutput=1

\PassOptionsToPackage{numbers, compress}{natbib}

\usepackage[final]{neurips_2024}

\usepackage[utf8]{inputenc} 
\usepackage[T1]{fontenc}    
\usepackage{hyperref}       
\usepackage{url}            
\usepackage{booktabs}       
\usepackage{amsfonts}       
\usepackage{nicefrac}       
\usepackage{microtype}      
\usepackage{xcolor}         
\usepackage{graphicx}
\usepackage{array}
\usepackage{amsmath}
\usepackage{mleftright}
\usepackage{multirow}
\usepackage{cleveref}
\usepackage{subfig}
\usepackage{makecell}
\usepackage{tikz}
\usepackage{scalerel}
\usepackage{fontawesome}
\usepackage{subcaption}

\usepackage[accsupp]{axessibility}  

\title{Estimating Ego-Body Pose from Doubly Sparse Egocentric Video Data}

\author{%
  Seunggeun Chi$^{\dagger}$ \\
  Purdue University\\
  \texttt{chi65@purdue.edu} \\
  \And
  Pin-Hao Huang$^{*}$, \ \  Enna Sachdeva\thanks{co-second authors} \\
  Honda Research Institute USA\\
  \texttt{\{pin-hao\_huang, enna\_sachdeva\}@honda-ri.com} \\
  \And
  Hengbo Ma\thanks{work done at Honda Research Institute USA} \\
  Honda Research Institute USA  \\
  \texttt{hengbo.academia@gmail.com} \\
  \And
  Karthik Ramani \\
  Purdue University\\
  \texttt{ramani@purdue.edu} \\
  \And
  Kwonjoon Lee \\
  Honda Research Institute USA\\
  \texttt{kwonjoon\_lee@honda-ri.com} \\
}

\def\rd{\textcolor{red}}

\def\gr{\textcolor{green}}
\def\bl{\textcolor{blue}}
\def\cy{\textcolor{cyan}}

\usepackage{fontawesome}

\newcommand\TalkingHead[1][]{\scalerel*{\begin{tikzpicture}[#1]
\useasboundingbox (-3.5,-4.5) rectangle (3.5,4.5);
\draw[xscale=-1,line width=pi*1mm] (-1,-4) to[out=50,in=-90,looseness=1.4] (-1,-3) to[out=-160,in=0]
   (-2,-3.2) to[out=180,in=-120] (-2.7,-2) to[out=120,in=-120] (-2.9,-1.5)
   to[out=120,in=-120] (-2.95,-1.1) to[out=60,in=-30] (-3.2,-0.9)
   to[out=150,in=-120] (-3,0.4) to[out=60,in=-90] (-3,1.2) 
   to[out=90,in=90,looseness=1.8] (3.3,1.2)
   to[out=-90,in=90,looseness=0.8] (2,-2)
   to[out=-90,in=150,looseness=0.8] (3.2,-4) -- cycle;
\end{tikzpicture}}{Q}}

\newcommand\HandPaper[1][]{\text{\faHandStopO}}

\begin{document}
\maketitle

\begin{abstract}
We study the problem of estimating the body movements of a camera wearer from egocentric videos. Current methods for ego-body pose estimation rely on \textit{temporally dense} sensor data, such as IMU measurements from \textit{spatially sparse} body parts like the head and hands. However, we propose that even \textit{temporally sparse} observations, such as hand poses captured intermittently from egocentric videos during natural or periodic hand movements, can effectively constrain overall body motion. Naively applying diffusion models to generate full-body pose from head pose and sparse hand pose leads to suboptimal results. To overcome this, we develop a two-stage approach that decomposes the problem into temporal completion and spatial completion. First, our method employs masked autoencoders to impute hand trajectories by leveraging the spatiotemporal correlations between the head pose sequence and intermittent hand poses, providing uncertainty estimates. Subsequently, we employ conditional diffusion models to generate plausible full-body motions based on these temporally dense trajectories of the head and hands, guided by the uncertainty estimates from the imputation. The effectiveness of our method was rigorously tested and validated through comprehensive experiments conducted on various HMD setup with AMASS and Ego-Exo4D datasets.

\end{abstract}
\section{Introduction}

\begin{figure}[!h]
    \centering
    \begin{minipage}{0.49\textwidth}
        \subfloat[]
        {\includegraphics[width=\linewidth, height=2cm]{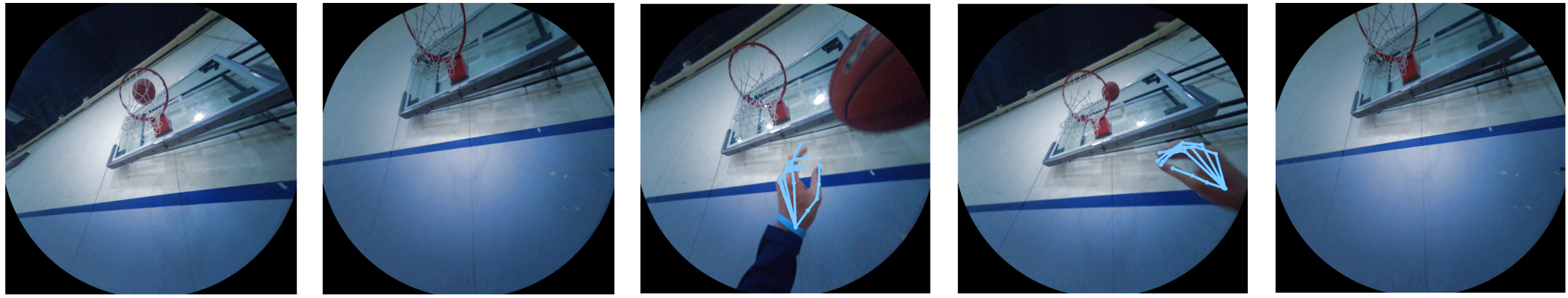}} \\
        \subfloat[]
        {\includegraphics[width=\linewidth, height=2.5cm]{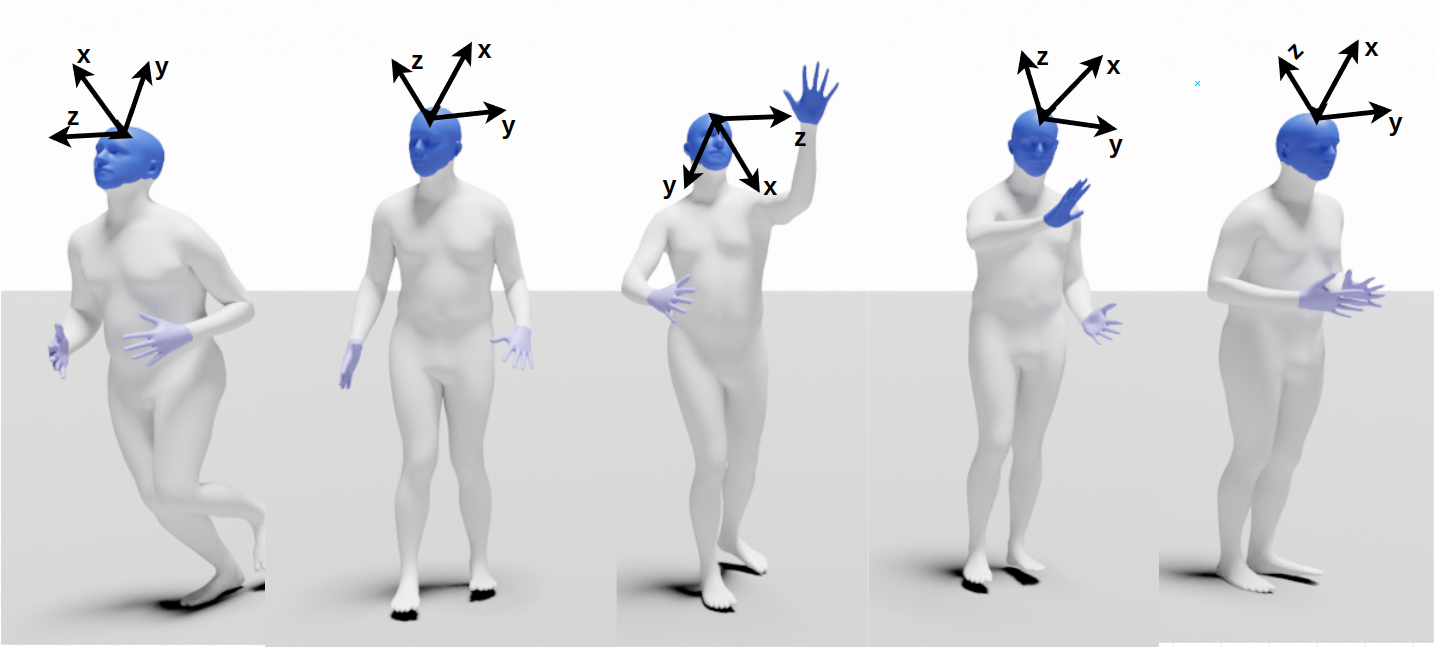}}
    \end{minipage}
    \hfill
    \begin{minipage}{0.49\textwidth}
       \subfloat[]
        {\includegraphics[width=\linewidth, height=2cm]{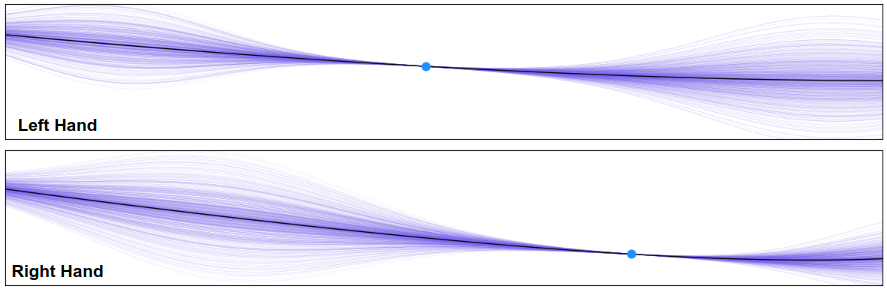}} \\
        \subfloat[]
        {\includegraphics[width=\linewidth, height=2.5cm]{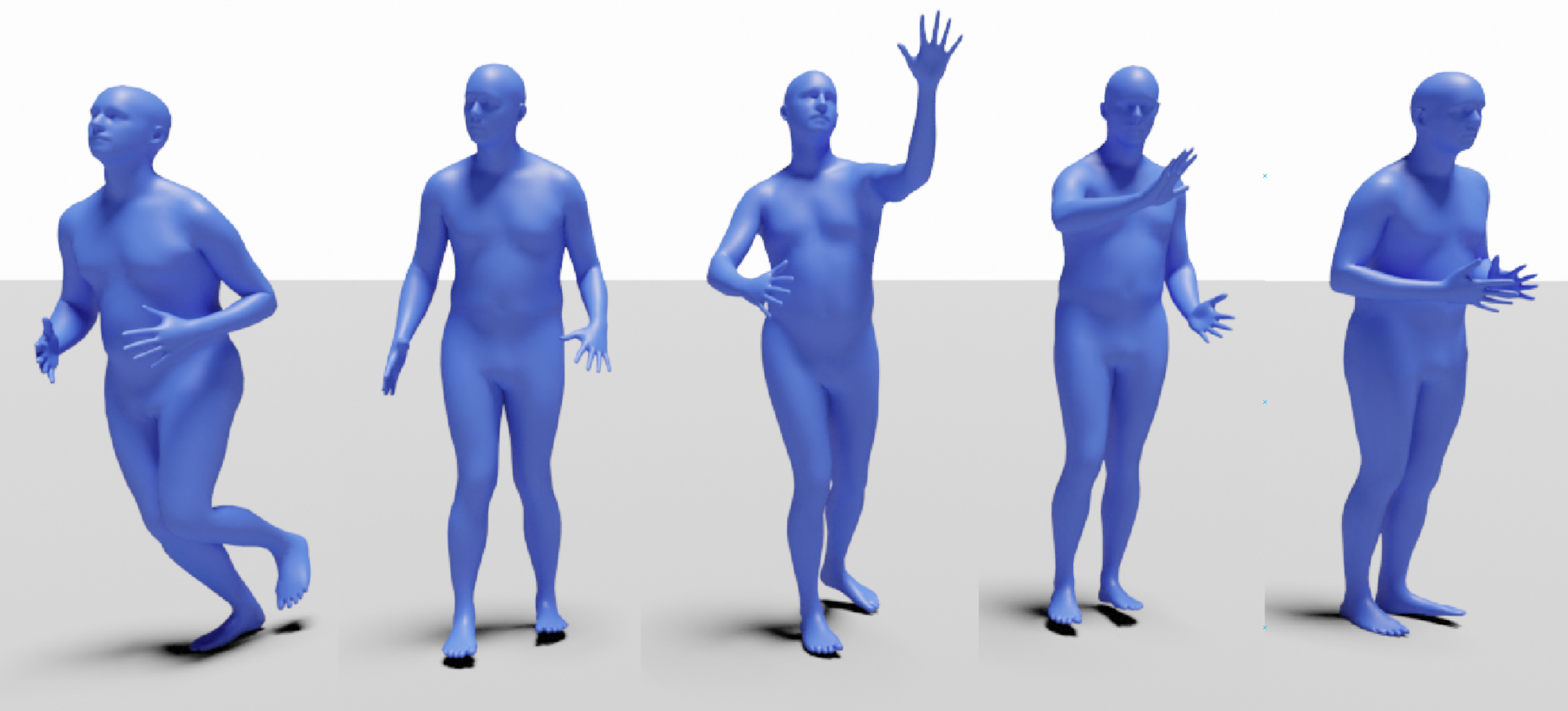}}
    \end{minipage}
     \caption{Overview of DSPoser. Our goal is to estimate ego-body pose without dependency on hand controllers in an HMD environment. (a) Given the egocentric video and head tracking signals as input, (b) our approach first predicts the hand pose in the frames where hands are visible \bl{(dark blue)}. It then estimates the hand poses in frames with invisible hands \cy{(light blue)} using imputation, and (c) estimates uncertainty associated with the hand poses where the hands are invisible, (d) The predicted and imputed hand pose is then used with head pose to predict the 3D full body pose.  } %
    \label{fig:teaser}
\end{figure}

The evolution of augmented reality (AR) devices such as the Apple Vision Pro, Meta Quest 3, Microsoft HoloLens 2, and etc. has dramatically reshaped interactive technologies. These head-mounted displays (HMDs) feature inertial measurement units (IMUs) and video capture capabilities, offering a unique egocentric perspective. However, their limited visibility of the user's body parts poses a significant challenge for accurate egocentric body pose estimation—a key element for immersive AR experiences.

Previous approaches have tackled this problem by spatially reconstructing the entire body from spatially sparse data. For instance, EgoEgo \cite{li2023ego} first estimates head poses using SLAM on the egocentric video, then generates body poses from these estimated head positions. Other methods, such as AvatarPoser \cite{jiang2022avatarposer} and BoDiffusion \cite{castillo2023bodiffusion}, primarily depend on temporally dense tracking signal from spatially sparse body parts, notably the head and hands. This dependency on specific hardware such as head-mounted displays and hand controllers constrains their versatility and diminishes their applicability in broader AR/VR scenarios where hand controllers might not be used, like sports training or analysis applications where the user needs to move freely without holding any devices, or augmented reality experiences in outdoor environments where carrying controllers is impractical. 

We observe that even \textit{temporally sparse} observations, such as hand poses captured intermittently from egocentric videos during natural or periodic hand movements, can effectively constrain overall body motion. While it is possible to utilize other visible body parts such as feet or elbows, we opted to rely on hand poses. This decision is based on the availability of hand pose detectors \cite{rong2021frankmocap, jiang2023rtmpose} and the fact that hands are visible in approximately 20$\%$ of video frames, as demonstrated in Table \ref{table:ratio}. Unlike previous work that concentrated only on spatial completion, our method incorporates temporal completion by leveraging the intermittent appearance of hands in egocentric videos. This dual completion approach not only enhances the robustness of body pose estimation under varying conditions but also reduces reliance on specific sensor hardware, making it more adaptable to various AR environments. In our setup, we use temporally sparse 3D hand poses from detections in egocentric videos combined with dense head tracking signals to reconstruct the full body. Initially, we temporally complete sparse hand information using a Masked Autoencoder (MAE) \cite{he2022masked}, which estimates hand pose trajectories by capturing the spatiotemporal correlations between intermittent hand poses and head tracking signals. We develop a probabilistic extension of the MAE to provide uncertainty estimates of the predicted hand pose sequence. Subsequently, using a conditional diffusion model, we spatially reconstruct the full body based on the head tracking signal data and imputed hand trajectories along with their predictive uncertainties. We call our approach \textbf{DSPoser} (Doubly Sparse Poser) because it can effectively utilize data that is \textit{doubly sparse} (sparse both temporally and spatially), as shown in Figure \ref{fig:teaser}.

This flexible framework is designed to seamlessly adapt to diverse AR/VR setups and devices, ranging from spatially sparse scenarios (e.g., using only head tracking signal or combining it with hand controllers) to doubly sparse scenarios (utilizing head signal data alongside hand detection from egocentric video). The key advantage lies in the assumption that the HMD's tracking signal is consistently available, enabling our approach to function across a wide range of environments and hardware configurations. Extensive experiments have proved our model's versatility and accurate pose estimation capabilities in various settings. Furthermore, our ablation studies highlight the significance of incorporating uncertainty estimates, as this crucial information enhances the overall quality of pose estimation, resulting in more reliable outputs. By addressing both temporal and spatial completion through our double completion approach, we have developed a robust and adaptable solution that reduces dependency on specific sensor hardware, making it well-suited for immersive AR experiences in diverse scenarios, such as sports training, outdoor environments, and beyond.
\newpage
In summary, our research presents three key contributions:
\begin{itemize}
\item A robust and versatile framework for egocentric body pose estimation tailored for HMDs. The framework adapts to various AR/VR settings and can leverage tracking signals available in most modern HMD devices without controllers.
\item We decomposed the problem into temporal completion and spatial completion. Our approach captures the uncertainty from hand trajectory imputation to guide the diffusion model for accurate full-body motion generation.
\item Extensive evaluations demonstrating the effectiveness of our framework on diverse datasets, outperforming existing methods and underscoring its potential for enhancing user interaction and immersion in AR experiences.
\end{itemize}

\section{Problem Formulation}
In our work, we aim to estimate the 3D human pose of a HMD user from sequences of RGB video and head tracking signal. We note that head tracking signal data is commonly accessible from IMU in most HMDs, such as Meta Quest and Apple Vision Pro. Suppose we are given an egocentric video $\mathcal{V}_\text{ego} = \{\mathcal{V}_1, \dots, \mathcal{V}_{T_w}\}$ where $\mathcal{V}_\tau$ is an RGB image and $T_w$ denotes the sequence length, and a corresponding head tracking signal sequence $\mathcal{T}_{\text{head}} = \{\mathcal{T}_1, \dots, \mathcal{T}_{T_w}\}$ where $\mathcal{T}_{\tau} \in \mathbb{R}^{D_{head}}$ and $D_{head}$ is the dimension of head tracking signal including 3D pose. Our goal is to estimate the full body pose $\mathcal{P} = \{\mathcal{P}_1, \dots, \mathcal{P}_{T_w}\}$, where pose state $\mathcal{P}_\tau \in \mathbb{R}^{J \times D}$ at time $\tau$, $J$ is the number of body joints and $D$ is the dimensionality of pose state.
We solve the problem of estimating $p(\mathcal{P}|\mathcal{V}_{ego},\mathcal{T}_{\text{head}})$ by decomposing it into 2 the stages of imputation and generation, assuming that we have \textit{temporally sparse} hand data $\ddddot{\mathcal{H}}$ from hand detection module $f(\cdot)$: $\ddddot{\mathcal{H}}=f(\mathcal{V}_{ego})$.
We first \textit{temporally complete} hand trajectory $\widetilde{\mathcal{H}}$ based on $\ddddot{\mathcal{H}}$ and $\mathcal{T}_{\text{head}}$, which can be written as $p(\widetilde{\mathcal{H}}|\ddddot{\mathcal{H}},\mathcal{T}_{\text{head}})$.
Then, we \textit{spatially complete} full body pose $\mathcal{P}$ from the imputed hands $\widetilde{\mathcal{H}}$ and $\mathcal{T}_{\text{head}}$, which can be written as  $p(\mathcal{P}|\widetilde{\mathcal{H}},\mathcal{T}_{\text{head}})$. Since $\widetilde{\mathcal{H}}$ is a probabilistic variable, we need to marginalize over $\widetilde{\mathcal{H}}$ as follows:
\begin{flalign}
\label{eq:marg}
p(\mathcal{P}|\mathcal{V}_{ego},\mathcal{T}_{\text{head}}) = 
\int_{\widetilde{\mathcal{H}}}p(\mathcal{P}|\widetilde{\mathcal{H}},\mathcal{T}_{\text{head}})p(\widetilde{\mathcal{H}}|f(\mathcal{V}_{ego}),\mathcal{T}_{\text{head}}).
\end{flalign}

\section{Methods} \label{sec:method}

\begin{figure}[h!]
    \vspace{-0.5em}
    \centering
    \includegraphics[width=1.0\textwidth]{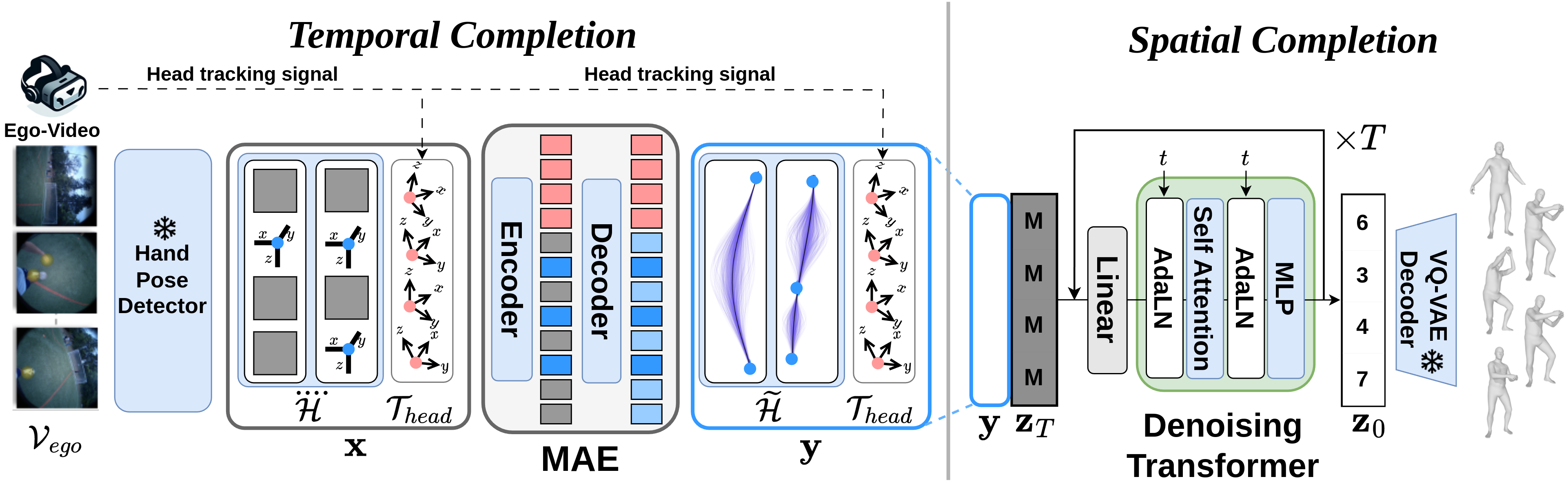}

    \caption{Overall pipeline of our proposed work DSPoser, composed of \textit{Temporal Completion} stage and \textit{Spatial Completion} stage to tackle pose estimation problem from \textit{doubly sparse} data.}
    \label{fig:pipeline}
    \vspace{-1.0em}
\end{figure}

\subsection{Detection: Hand Pose Estimation from Egocentric Video}

In this work, we estimate the 3D position of the hand from an egocentric camera using a two-step process. First, we use FrankMocap \cite{rong2021frankmocap} to predict hand poses as SMPL-X parameters \cite{pavlakos2019expressive}, from which we extract local 3D hand joint positions relative to the root of the hand model's kinematic tree, denoted as $\mathcal{H}^{3D}_{h} \in \mathbb{R}^{21 \times 3}$. Simultaneously, we use RTM-Pose \cite{jiang2023rtmpose} to estimate 2D hand joint positions in the image, $\mathcal{H}^{2D}_I \in \mathbb{R}^{21 \times 2}$.  
Finally, we determine the 3D hand joint positions in the camera coordinate system, $\mathcal{H}^{3D}_I = \mathcal{H}^{3D}_{h} + \mathbf{d}$ by solving for $\mathbf{d} \in \mathbb{R}^3$ that minimizes the reprojection error $\| \mathcal{H}^{2D}_I - \mathbf{K}(\mathcal{H}^{3D}_{h} + \mathbf{d}) \|_2$. Here, $\mathbf{K}$ is the intrinsic matrix, obtained by transforming the original camera parameters into a pinhole model through undistortion.

To better constrain the hand trajectories, we attempted to obtain rotation information from the 3D hand detection. However, due to the inconsistent quality of hand detection, the rotational information derived from the hand pose was noisy. Therefore, we decided not to incorporate this rotational information into our hand tracking approach on the Ego-Exo4D dataset. We utilized only the 3D wrist location from the Ego-Exo4D dataset, represented by $D_{hand} = 3$. In contrast, for the AMASS dataset, we leveraged both rotational information and 3D location, as this data is readily available, resulting in $D_{hand} = 9$.

\subsection{Temporal Completion: Hand Trajectory Imputation from Sparse Hand Pose}

\paragraph{Masked Auto-Encoder (MAE)}

In our work, we employed a Masked Autoencoder (MAE) \cite{he2022masked} to impute missing hand trajectories using head tracking signal $\mathcal{T}_{head}$ and detected hand pose $\ddddot{\mathcal{H}}$. Inspired by Vision Transformer (ViT), we treated each $\mathcal{T}_\tau$ and $\ddddot{\mathcal{H}}_\tau$ at time $\tau$ as a token similar to an image patch in ViT. To accommodate this, we implemented two embedding layers, one for head tracking signal $\mathcal{T}_{\tau} \in \mathbb{R}^{D_{head}}$ and the other for hand $\ddddot{\mathcal{H}}_\tau \in \mathbb{R}^{D_{hand}}$, both projecting into the common token dimension $D_M$. For the AMASS dataset, we follow the head tracking signal representation $D_{\text{head}} = 18$ as in \cite{jiang2022avatarposer}. For the Ego-Exo4D dataset, $D_{\text{head}} = 15$, which includes head position and left/right IMU signals. Consequently, the total number of token amounts to $3 \times T_w$, where 3 accounts for the head and both hands, and $T_w$ is the sequence length. Sinusoidal positional encoding (PE) is used for both the encoder and decoder patches after tests showed it suffices for learning different modalities, compared to learnable PE. In an HMD environment, we assume that the head tracking signal $\mathcal{T}_{head}$ is always available, but hand visibility depends on the egocentric video. Thus, masking is applied only to the hand tokens based on their visibilities within egocentric view. 

In contrast to the MAE \cite{he2022masked} training approach, which maintains a consistent number of masked patches due to a fixed masking ratio, the count of frames with invisible hand varies across instances in our setup.
To address this variability, our encoder selectively applies attention masking to these inputs, ensuring that queries do not attend to tokens where hand is invisible.
This attention masking technique adapts dynamically to the fluctuating numbers of missing frames across the instances, enhancing the model's ability to handle data sparsity effectively.
For decoder, we adopted MAE decoder design except the last projection layer to guide the uncertainty. To capture the uncertainty, we split the final projection layer into two heads for mean and variance of a Gaussian distribution.

\paragraph{Uncertainty-aware MAE}
Following the  \cite{seitzer2022pitfalls, valdenegro2022deeper}, to make the MAE aware of the predictive uncertainty of imputed hand pose sequence, we employ the $\beta$-NLL loss \cite{seitzer2022pitfalls} function to manage uncertainty by using a set of mean heads $\mathbf{\mu}_i (\mathbf{x})$ and variance heads $\mathbf{\sigma}_i^2 (\mathbf{x})$, which are derived from $M$ models initialized differently, where $\mathbf{x}=[\ddddot{\mathcal{H}};\mathcal{T}]$ is an input to the MAE and $i \in [1, M]$. The mean heads $\mathbf{\mu}_i (\mathbf{x})$ and variance heads $\mathbf{\sigma}_i^2(\mathbf{x})$ are trained using the Gaussian negative log-likelihood loss, which applies to each sample indexed by $n$ with input $\mathbf{x}_n$ and ground truth hand pose sequence $\mathbf{y}_n$.
\begin{flalign}
L_{\beta-\text{NLL}}(\mathbf{y}_n, \mathbf{x}_n) &= \text{sg}(\sigma_i^{2\beta}) L_{\text{NLL}}(\mathbf{y}_n, \mathbf{x}_n) \text{ where},\\
L_{\text{NLL}}(\mathbf{y}_n, \mathbf{x}_n) &= \frac{\log \sigma_i^2(\mathbf{x}_n)}{2} + \frac{(\mu_i(\mathbf{x}_n) - \mathbf{y}_n)^2}{2\sigma_i^2(\mathbf{x}_n)}.
\end{flalign}
The $L_{\text{NLL}}$ loss function causes the predicted variance to act as a weighting factor for each data point, emphasizing those with higher variances. The parameter $\beta$ adjusts the intensity of this weighting. The $\text{sg($\cdot$)}$ function is used to apply the stop-gradient operation, thus preventing gradients from propagating through this part of the computation.

After training, we measure the aleatoric (data) uncertainty $\mathcal{U}_{ale}(\cdot)$ by averaging the variances across models, and epistemic (model) uncertainty $\mathcal{U}_{epi}(\cdot)$ by calculating variance of model means, and total uncertainty by adding both uncertainties:
\begin{flalign}
\mathcal{U}_{ale}(\mathbf{x}) &= \mathbb{E}_i[\sigma_i^{2}(\mathbf{x})] \approx M^{-1} \sum_i \sigma_i^2 (\mathbf{x})\label{eq:ale}\\
\mathcal{U}_{epi}(\mathbf{x}) &= \text{Var}_i[\mu_i(\mathbf{x})] \label{eq:epi} \\ 
\mathcal{U}_{tot}(\mathbf{x}) &= \mathcal{U}_{ale}(\mathbf{x}) + \mathcal{U}_{epi}(\mathbf{x}) \label{eq:tot}
\end{flalign}
Note that $\mathcal{U}_{ale}(\cdot)$ and $\mathcal{U}_{epi}(\cdot)$ provide uncertainties for each frame and each pose state dimension. The captured uncertainty is visualized in Figure \ref{fig:uncertainty}, demonstrating that MAE effectively captures uncertainty.

\subsection{Spatial Completion: Uncertainty-guided Body Pose Generation from Imputed Hand Trajectories and Head Tracking Signal}

We employed the VQ-Diffusion \cite{rombach2022high} to generate full body poses from imputed hand trajectories and head tracking signal. The exposition of VQ-Diffusion can be found in Section \ref{sec:prelim} of the Appendix.
As illustrated in \Cref{fig:pipeline}, our motion generation module is designed to generate human motion sequences from the temporally dense hand and head trajectories with uncertainty obtained from the MAE model.

\paragraph{VQ-VAE}
We first train the VQ-VAE to represent human motion with a discrete codebook representation as described in \Cref{sec:vqvae}. 
We mostly followed the architectural design and training methods of \cite{zhang2023generating}.
After the codebook representation is learned by the VQ-VAE, we utilize this latent codebook representation to train a denoising diffusion model.

\paragraph{Denoising Transformer}
Motivated by the work of VQ-Diffusion, we design a denoising transformer that estimates the distribution $p(\mathbf{z}_0|\mathbf{z}_t,\mathbf{y})$. An overview of our proposed model is depicted in \Cref{fig:pipeline}. We closely follow the implementation of \cite{chi2024m2d2m}. To incorporate the diffusion step $t$ into the network, we employ the adaptive layer normalization (AdaLN) \cite{ba2016layer,lee2022vitgan}. We concatenated the estimated hand and head trajectory with codebook after a embedding layer, to match the dimension with codebook representation. Finally, we use the decoder to decode $\mathbf{z}_0$ to obtain a full body pose sequence.  
\paragraph{Uncertainty Guidance}
We introduce several strategies to guide the denoising process using uncertainty estimates of imputed hand trajectories: sampling, dropout, and distribution embedding. 

For \textit{sampling}, we sample a hand sequence from the distribution $\widetilde{\mathcal{H}}\sim\mathcal{N}(\mu^{*}(\mathbf{x}), \sqrt{\mathcal{U^*}(\mathbf{x})})$ and regard it as the conditioning vector $\mathbf{y}$, where $\mu^*(\mathbf{x}) = \mathbb{E}_i[\mu_i(\mathbf{x})] \approx M^{-1} \sum_i \mu_i(\mathbf{x})$ and $\mathcal{U^*}(\mathbf{x})$ is measured by one of Eq. (9), (10), and (11). While it would be ideal to sample multiple times to better approximate the marginalization in Equation \ref{eq:marg}, we find just using one sample provides a competitive performance.

For \textit{dropout}, we set each dimension of $\mu^{}(\mathbf{x})$ to zero with a certain probability, which is determined by the corresponding dimension of $\mathcal{U^*}(\mathbf{x})$, and denote the result as $y$.
The probability of the $d$-th dimension of $\mu(\mathbf{x})$ being zero is $p_d = 1 - (\mathcal{U}_d^*(\mathbf{x}) - \mathcal{U}_{d^{min}}^*(\mathbf{x})) / (\mathcal{U}_{d^{max}}^*(\mathbf{x}) - \mathcal{U}_{d^{min}}^*(\mathbf{x}))$ where $\mathcal{U}_d^*(\mathbf{x})$ is the $d$-th dimension of $\mathcal{U}^*(\mathbf{x})$, and $\mathcal{U}_{d^{min}}^*(\mathbf{x})$, $\mathcal{U}_{d^{max}}^*(\mathbf{x})$ are the minimum and maximum values over the sequence length, respectively.

For \textit{distribution embedding} \cite{hilbertembeddings}, we embed the Gaussian distribution $\mathcal{N}(\mu^{*}(\mathbf{x}), \sqrt{\mathcal{U^*}(\mathbf{x})})$ to a vector by concatenating the $\mu^*(\mathbf{x})$ and $\mathcal{U}^*(\mathbf{x})$ in the feature dimension. The resulting embedding will be further concatenated with the head pose sequence to form a conditioning vector $\mathbf{y}$.

\section{Experiments} \label{sec:exp}

\subsection{Datasets $\&$ Evaluation Metrics} \label{sec:dataset}

\paragraph{Ego-Exo4D dataset}
Ego-Exo4D \cite{grauman2023ego} contains simultaneous captures of egocentric (first-person) and exocentric (third-person) video perspectives of participants performing complex activities like sports, dance, and mechanical tasks. The dataset comprises 1,422 hours of video ranging from 1 to 42 minutes per video. In addition to video, it provides camera poses, IMU data, and human pose annotations. Specifically for the egopose task, it includes separate training and validation video sets containing 334 and 83 videos respectively. Our problem formulation of ego body pose estimation differs from the ego body pose prediction task from \cite{grauman2023ego}, which aims to predict a single future frame given a specific time window.
\paragraph{AMASS dataset }
The AMASS dataset \cite{mahmood2019amass} is a large human motion database that unifies different existing optical marker-based MoCap datasets by converting them into realistic 3D human meshes represented by SMPL \cite{loper2023smpl} model parameters. Following the AvatarPoser \cite{jiang2022avatarposer} evaluation, we used the CMU \cite{CMUGraphicsLab2000}, BMLrub \cite{troje2002decomposing}, and HDM05 \cite{muller2007documentation} subsets from the AMASS dataset and their preprocessing of tracking signal information.
Since AMASS does not include RGB images, we set $D_{hand} = 9$ assuming that 3D hand position and 6D rotation is available when the hand is "visible". To determine visibility, we compute the angle between the z-axis vector of the head rotation and the vector from the head position to the hand. We define the hand as "visible" if this angle is within a 45° range, corresponding to a 90° field of view (FoV) of HMD devices.
\paragraph{Evaluation metric} We evaluate our results using the following metrics: Mean Per Joint Position Error (MPJPE), Mean Per Joint Velocity Error (MPJVE), and Mean Per Joint Rotation Error (MPJRE), following the evaluation of \cite{jiang2022avatarposer,castillo2023bodiffusion}. Since Ego-Exo4D dataset doesn't have the annotations for 6D rotation, MPJRE is reported only for AMASS. We report all values with the confidence interval of 95\%. We also provide details on MPJPE across hands, upper body above the pelvis, and lower body below the pelvis, denoted as Hand PE, Upper PE, and Lower PE, respectively.

\subsection{Full Body Pose Estimation from Doubly Sparse data}
\label{section:exp_doubly_sparse}

\begin{table}[t]
\centering
\caption{Performance comparisons across baseline models for doubly sparse video data on the \textbf{AMASS} test set. We report MPJRE [°], MPJPE [cm], and MPJVE [cm/s], with the best results highlighted in \textbf{boldface}. Models trained by us are marked with $^{*}$. The notation $\ddddot{\text{data}}$ denotes temporally sparse data, $\widetilde{\text{data}}$ indicates imputed data, and all other cases involve dense data. $T_s$ indicates the sliding window, $\textbf{x}$ indicates the input of our whole pipeline, and $\textbf{y}$ indicates the input of denoising Transformer.} \label{table:amass_double}
\resizebox{\textwidth}{!}{
\begin{tabular}{lccccccc}
\hline
Methods  & $T_s$ & $\textbf{x}$ & Imputation & $\textbf{y}$ & MPJPE & MPJVE & MPJRE \\\hline \hline
VQ-VAE (Recons) & 20 & Full body & - & Full body & 1.26 & 11.37 & 1.81 \\\hline
$\text{EgoEgo}^*$ \cite{li2023ego} & 20 & $\large\TalkingHead$ & - & $\large\TalkingHead$ & 19.17 & 46.17 & 7.30 \\
$\text{Bodiffusion}^*$ \cite{castillo2023bodiffusion} & 20 & $\large\TalkingHead$ & - & $\large\TalkingHead$ & 19.27 & 60.29 & 8.51 \\
\textbf{DSPoser (Ours)} & 20 & $\large\TalkingHead$ & - & $\large\TalkingHead$ & $12.08^{\pm{0.04}}$ & $75.07^{\pm{0.26}}$ & $7.04^{\pm{0.02}}$\\ 
\textbf{DSPoser (Ours)} & 20 & $\large\TalkingHead$ & MAE & $\large\TalkingHead$ \& $\widetilde{\HandPaper}$ & $\textbf{7.06}^{\pm{0.02}}$ & $\textbf{28.26}^{\pm{0.05}}$ & $\textbf{5.00}^{\pm{0.01}}$ \\ \hline
Bodiffusion \cite{castillo2023bodiffusion} & 20 & $\large\TalkingHead$ \& $\ddddot{\HandPaper}$ & Interpolation & $\large\TalkingHead$ \& $\widetilde{\HandPaper}$ & 46.45 & 75.33 & 17.99 \\
Bodiffusion \cite{castillo2023bodiffusion} & 20 & $\large\TalkingHead$ \& $\ddddot{\HandPaper}$ & MAE & $\large\TalkingHead$ \& $\widetilde{\HandPaper}$ & 7.35 & 31.33 & 5.47 \\
\textbf{DSPoser (Ours)} & 20 & $\large\TalkingHead$ \& $\ddddot{\HandPaper}$ & MAE & $\large\TalkingHead$ \& $\widetilde{\HandPaper}$ & $\textbf{5.51}^{\pm{0.02}}$ & $\textbf{24.19}^{\pm{0.10}}$ & $\textbf{4.09}^{\pm{0.02}}$ \\ \hline
AvatarPoser \cite{jiang2022avatarposer} & 1 & $\large\TalkingHead$ \& $\ddddot{\HandPaper}$ & Interpolation & $\large\TalkingHead$ \& $\widetilde{\HandPaper}$ & 40.42 & 64.07 & 16.37 \\
AvatarJLM \cite{zheng2023realistic} & 1 & $\large\TalkingHead$ \& $\ddddot{\HandPaper}$ & Interpolation & $\large\TalkingHead$ \& $\widetilde{\HandPaper}$ & 25.02 & 68.42 & 14.14 \\
AvatarPoser \cite{jiang2022avatarposer} & 1 & $\large\TalkingHead$ \& $\ddddot{\HandPaper}$ & MAE & $\large\TalkingHead$ \& $\widetilde{\HandPaper}$ & 9.88 & 62.31 & 5.98 \\
AvatarJLM \cite{zheng2023realistic} & 1 & $\large\TalkingHead$ \& $\ddddot{\HandPaper}$ & MAE & $\large\TalkingHead$ \& $\widetilde{\HandPaper}$ & 7.12 & \textbf{37.60} & 5.24 \\
\textbf{DSPoser (Ours)} & 1 & $\large\TalkingHead$ \& $\ddddot{\HandPaper}$ & MAE & $\large\TalkingHead$ \& $\widetilde{\HandPaper}$ & $\textbf{5.87}^{\pm{0.13}}$ & $49.12^{\pm{0.24}}$ & $\textbf{4.31}^{\pm{0.10}}$ \\ \hline
\end{tabular}
}
\end{table}

\begin{table}[t]
\centering
\caption{Performance comparisons across baseline models for doubly sparse video data on the \textbf{Ego-Exo4D} validation set. We report MPJPE [cm] and MPJVE [cm/s], with the best results highlighted in \textbf{boldface}. Models trained by us are marked with $^{*}$. The notation $\ddddot{\text{Data}}$ denotes temporally sparse data, $\widetilde{\text{data}}$ indicates imputed data, and all other cases involve dense data.
} \label{table:egoexo_double}
\resizebox{0.90\textwidth}{!}{
\begin{tabular}{lcccccc}
\hline
Methods  & $T_s$ & $\textbf{x}$ & Imputation & $\textbf{y}$ & MPJPE & MPJVE \\ \hline \hline
VQ-VAE (Recons) & 20 & Full body & - & Full body & 6.77 & 33.29 \\\hline
$\text{EgoEgo}^*$ \cite{li2023ego} & 20 & $\large\TalkingHead$ & - & $\large\TalkingHead$  & 29.49 & 47.50 \\
$\text{Bodiffusion}^*$ \cite{castillo2023bodiffusion} & 20 & $\large\TalkingHead$ & - & $\large\TalkingHead$  & 28.56 & 109.71 \\ 
\textbf{DSPoser (Ours)} & 20 & $\large\TalkingHead$ & - & $\large\TalkingHead$ & $19.12^{\pm{0.06}}$ & $48.54^{\pm{0.11}}$ \\
\textbf{DSPoser (Ours)} & 20 & $\large\TalkingHead$ & MAE & $\large\TalkingHead$ \& $\widetilde{\HandPaper}$ & $\textbf{18.46}^{\pm{0.06}}$ & $\textbf{40.67}^{\pm{0.11}}$ \\
\hline
$\text{Bodiffusion}^*$ \cite{castillo2023bodiffusion} & 20 & $\large\TalkingHead$ \& $\ddddot{\HandPaper}$ & Interpolation & $\large\TalkingHead$ \& $\widetilde{\HandPaper}$ & 59.81 & 120.12\\
$\text{Bodiffusion}^*$ \cite{zheng2023realistic} & 20 & $\large\TalkingHead$ \& $\ddddot{\HandPaper}$ & MAE & $\large\TalkingHead$ \& $\widetilde{\HandPaper}$ & 22.12     & 53.30 \\
\textbf{DSPoser (Ours)} & 20 & $\large\TalkingHead$ \& $\ddddot{\HandPaper}$ & MAE & $\large\TalkingHead$ \& $\widetilde{\HandPaper}$ & $\textbf{16.84}^{\pm{0.04}}$ & $\textbf{39.86}^{\pm{0.05}}$  \\\hline
$\text{AvatarPoser}^*$ \cite{jiang2022avatarposer} & 1 & $\large\TalkingHead$ \& $\ddddot{\HandPaper}$ & Interpolation & $\large\TalkingHead$ \& $\widetilde{\HandPaper}$ & 47.28 & 89.34 \\
$\text{AvatarJLM}^*$ \cite{zheng2023realistic} & 1 & $\large\TalkingHead$ \& $\ddddot{\HandPaper}$ & Interpolation & $\large\TalkingHead$ \& $\widetilde{\HandPaper}$ & 43.01 & 61.98  \\
$\text{AvatarPoser}^*$ \cite{jiang2022avatarposer} & 1 & $\large\TalkingHead$ \& $\ddddot{\HandPaper}$ & MAE & $\large\TalkingHead$ \& $\widetilde{\HandPaper}$ & 24.54 & 62.34  \\
$\text{AvatarJLM}^*$ \cite{castillo2023bodiffusion} & 1 & $\large\TalkingHead$ \& $\ddddot{\HandPaper}$ & MAE & $\large\TalkingHead$ \& $\widetilde{\HandPaper}$ & 21.08 & \textbf{45.77} \\
\textbf{DSPoser (Ours)} & 1 & $\large\TalkingHead$ \& $\ddddot{\HandPaper}$ & MAE & $\large\TalkingHead$ \& $\widetilde{\HandPaper}$ & $\textbf{19.09}^{\pm{0.21}}$ & $55.82^{\pm{0.27}}$  \\\hline
\end{tabular}
}
\end{table}

To demonstrate the effectiveness of our framework on doubly sparse egocentric video data, we investigated the results of our framework, DSPoser, on the AMASS dataset and Ego-Exo4D, as shown in \Cref{table:amass_double} and \Cref{table:egoexo_double}, respectively.
Since the task of body pose estimation from doubly sparse data is newly introduced in our paper, we compare our results to other baselines, EgoEgo \cite{li2023ego}, Bodiffusion \cite{castillo2023bodiffusion}, AvatarPoser \cite{jiang2022avatarposer}, and AvatarJLM \cite{castillo2023bodiffusion}.
Those baselines are designed to estimate human body poses from spatially sparse data. EgoEgo estimates body poses from head poses, and the others estimate body poses from head and hand tracking signals. We report the experimental results using the sampling strategy with aleatoric uncertainty unless otherwise stated.
To train the baslines on temporally sparse data, we extend the algorithm as follows: (1) \textit{Interpolation}: we imputed hand poses with linear interpolation; (2) \textit{MAE}: we use our trained MAE to impute the hand trajectory. In $T_s = 1$ setup, we report our result after averaging 16 samples while the result in $T_s = 20$ setup is from a single sample.

As shown in \Cref{table:amass_double}, DSPoser consistently outperforms baseline methods on AMASS across all metrics, underscoring the effectiveness of our two-stage approach for ego-body pose estimation. DSPoser achieves notable improvements in MPJPE for both sliding window sizes, $T_s = 20$ and $T_s = 1$. For $T_s = 20$, DSPoser reduces MPJPE from 7.35 cm to 5.51 cm, significantly outperforming the Bodiffusion extension, which uses MAE to impute invisible hands. For $T_s = 1$, DSPoser achieves superior MPJPE compared to AvatarJLM, though it showswlimitations in MPJVE due to the stochasticity of the diffusion model. 
In the experimental results presented in \Cref{table:egoexo_double}, our DSPoser model demonstrates superior performance on the Ego-Exo4D validation set. The model outperforms existing baselines, achieving a lower MPJPE of 16.84 cm, which represents an improvement over the next best model by 5.49 cm. Additionally, DSPoser achieves an MPJVE of 39.86 cm/s, improving upon the basline of naive extension of Bodiffusion by 7.64 cm/s. 

It is evident that by incorporating temporally sparse hand pose data, our DSPoser framework significantly enhances pose estimation accuracy. For instance, on the AMASS dataset, MPJPE improved dramatically from 12.08 cm to 5.51 cm, while on the Ego-Exo4D dataset, it improves from 19.12 cm to 16.84 cm in $T_s = 20$ setup. This indicates that even sparse hand trajectory data, when effectively utilized, can provide crucial information for refining the accuracy of ego body pose estimation. Our method's ability to harness sparsely available data underscores its potential in applications where capturing dense sequence is challenging.

\begin{table}[t]
\centering
\caption{Performance comparisons across baseline models on the \textbf{AMASS} test set. We report MPJRE [°], MPJPE [cm], and MPJVE [cm/s], with the best results highlighted in \textbf{boldface}.
Note that $^{\ddagger}$ is trained only with dense data without uncertainty.}\label{table:amass_spatial}
\resizebox{\textwidth}{!}{
\begin{tabular}{lccccccc}
\hline
Methods & $\textbf{y}$ & MPJPE & MPJVE & MPJRE & Hand PE & Upper PE & Lower PE \\ \hline \hline
FinalIK \cite{RootMotionFinalIK2018} & $\large\TalkingHead$ \& $\HandPaper$ & 18.09 & 59.24 & 16.77 & - & - & -\\
LoBSTR \cite{yang2021lobstr} & $\large\TalkingHead$ \& $\HandPaper$ & 9.02 & 44.97 & 10.69 & - & - & -\\
VAE-HMD \cite{dittadi2021full} & $\large\TalkingHead$ \& $\HandPaper$  & 6.83 & 37.99 & 4.11& - & - & -\\
CollMoves \cite{ahuja2021coolmoves} & $\large\TalkingHead$ \& $\HandPaper$ & 5.55 & 65.28 & 4.58 & - & - & -\\
AvatarPoser \cite{jiang2022avatarposer} & $\large\TalkingHead$ \& $\HandPaper$ & 4.20 & 28.23 & 3.08 & 2.34 & 1.88 & 8.06 \\
AvatarJLM \cite{zheng2023realistic} & $\large\TalkingHead$ \& $\HandPaper$ & \textbf{3.35} & \textbf{20.79} & \textbf{2.90} & \textbf{1.24} & \textbf{1.72} & \textbf{6.20} \\
$\textbf{DSPoser (Ours)}^{\ddagger}$ & $\large\TalkingHead$ \& $\HandPaper$ & $3.73^{\pm{0.08}}$ & $43.43^{\pm{0.14}}$ & $2.94^{\pm{0.09}}$ & 3.26 & 1.92 & 6.53 \\ \hline
\end{tabular}
}
\end{table}

\subsection{Full Body Pose Estimation from Spatially Sparse data}

To demonstrate the versatility of our framework, we conduct experiments on spatially sparse video data. In the temporally dense data setup, where there is no uncertainty regarding hand poses, the dense data directly works as a condition $\mathbf{y}$ for spatial completion on the right side of \Cref{fig:pipeline}.
\Cref{table:amass_spatial} presents the results, demonstrating that DSPoser performs comparably to baseline models designed specifically for dense data setups on MPJPE and MPJRE metrics, underscoring the versatility of our dual approach in handling dense data scenarios. As discussed in \Cref{section:exp_doubly_sparse}, the higher MPJVE error results from the inherent stochasticity of the diffusion model.

\subsection{Ablation Studies}

\begin{table}[t]
\begin{minipage}{0.49\textwidth}
    \centering
    \caption{Ablation study for uncertainty guidance strategy}\label{table:uncertainty_strategy}
    \resizebox{\textwidth}{!}{
    \begin{tabular}{lccc}
        \hline
        Methods & MPJPE & MPJVE & MPJRE \\ \hline \hline
        w/o Uncertainty & $6.05^{\pm{0.01}}$ & $30.12^{\pm{0.04}}$ & $4.36^{\pm{0.00}}$ \\
        Sample & $\textbf{5.51}^{\pm{0.02}}$  & $\textbf{24.19}^{\pm{0.10}}$ & $\textbf{4.09}^{\pm{0.02}}$ \\
        Distribution emb. & $5.67^{\pm{0.02}}$ & $25.63^{\pm{0.02}}$ & $4.16^{\pm{0.02}}$ \\
        Dropout & $5.55^{\pm{0.02}}$ & $25.10^{\pm{0.02}}$ & $4.11^{\pm{0.02}}$ \\ \hline
    \end{tabular}
    }
\end{minipage}
\hfill
\begin{minipage}{0.48\textwidth}
    \centering
    \caption{Ablation study for different types uncertainty}\label{table:uncertainty_types}
    \resizebox{\textwidth}{!}{
    \begin{tabular}{lccc}
        \hline
        Methods & MPJPE & MPJVE & MPJRE \\ \hline \hline
        w/o Uncertainty & $6.05^{\pm{0.01}}$ & $30.12^{\pm{0.04}}$ & $4.36^{\pm{0.00}}$ \\
        Epistemic & $5.78^{\pm{0.03}}$ &$ 27.55^{\pm{0.09}} $& $4.10^{\pm{0.02}}$ \\
        Aleatoric & $\textbf{5.51}^{\pm{0.02}}$  & $\textbf{24.19}^{\pm{0.10}}$ & $\textbf{4.09}^{\pm{0.02}}$ \\
        Total & $5.59^{\pm{0.02}}$ & $25.65^{\pm{0.09}}$ &$ 4.16^{\pm{0.02}}$ \\ \hline
    \end{tabular}
    }
\end{minipage}
\end{table}

\begin{table}[t!]
    \centering
    \begin{minipage}{0.33\textwidth}
        \centering
        \caption{Ablation study for $\beta$ for uncertainty capturing with MAE.}\label{table:betanll}
        \resizebox{0.70\textwidth}{!}{
        \begin{tabular}{cc}
            \hline
            $\beta$ & MPJPE (cm) \\ \hline \hline
            1.00      &  11.57 \\
            0.50      &  10.85 \\
            0.25      &  10.92 \\\hline
        \end{tabular}
        }
    \end{minipage}
    \hfill
    \begin{minipage}{0.29\textwidth}
        \centering
        \caption{Hand detection accuracy on Ego-Exo4D dataset.}\label{table:detector_error}
        \resizebox{0.83\textwidth}{!}{
            \begin{tabular}{c|c}
                \hline
                & MPJPE (cm) \\
                \hline
                Left Hand & 9.51 \\
                Right Hand & 9.63 \\
                \hline
            \end{tabular}
        }
    \end{minipage}
    \hfill
    \begin{minipage}{0.35\textwidth}
        \centering
        \caption{Hand visibility ratio for AMASS and Ego-Exo4D dataset}\label{table:ratio}
        \resizebox{\textwidth}{!}{
    \begin{tabular}{cccc}
        \hline
        Num. Hands & 0 & 1 & 2 \\
        \hline
        AMASS               & 82.77\% & 12.04\% & 5.19\%\\ 
        Ego-Exo4D              & 72.95\% & 19.44\% &  7.61\%\\ 
        \hline

    \end{tabular}
        }
    \end{minipage}
\end{table}

Based on the ablation study results shown in Tables \ref{table:uncertainty_strategy} and \ref{table:uncertainty_types}, we can analyze the impact of different uncertainty guidance strategies and types of uncertainty on the performance of the model for body pose estimation. The ablation study is conducted with AMASS dataset with the sliding window $T_s = 20$ to better analyze the effect of the uncertainty guidance.
\Cref{table:uncertainty_strategy} investigates the effects of various uncertainty guidance strategies, including no uncertainty guidance, sample, distribution embedding, and dropout. The results suggest that incorporating uncertainty guidance through these strategies can improve the model's performance across different metrics. The sampling strategy achieves the best performance, with the lowest MPJPE of 5.51, MPJVE of 24.19, and MPJRE of 4.09, indicating its effectiveness in capturing uncertainty and improving pose estimation accuracy.

\Cref{table:uncertainty_types} examines the contributions of different types of uncertainty, including epistemic uncertainty, aleatoric uncertainty, and total uncertainty. The results show that accounting for aleatoric uncertainty leads to the best overall performance. This suggests that considering data uncertainty can provide complementary information and improve the robustness of the pose estimation model.
Overall, the ablation study highlights the importance of incorporating uncertainty guidance and considering different types of uncertainty in the model design for accurate and reliable body pose estimation.

In \Cref{table:betanll}, we analyzed the effect of different $\beta$ values on the AMASS dataset during the uncertainty capturing process of the Masked Auto-Encoder (MAE). The results, shown in the table, indicate that $\beta = 0.5$ provides the best temporal completion for head and hand 3D positions from the doubly sparse input. Therefore, we set $\beta$ to 0.5 for training the MAE.

\subsection{Hand Detection Accuracy and Hand Visibility Statistics}
We investigate the error of the hand detector applied to the Ego-Exo4D dataset in terms of MPJPE, as shown in \Cref{table:detector_error}. The detection results indicate an average error of less than 10 cm. We also analyze the visibility statistics for the AMASS and Ego-Exo4D datasets in \Cref{table:ratio}. In the AMASS dataset, at least one hand is visible in 18\% of all frames with a 90° field of view (FoV), whereas in the Ego-Exo4D dataset, at least one hand is visible in 27\% of all frames.

\subsection{Qualitative Results}
\begin{figure}[t]
    \centering
    \includegraphics[width=1.0\textwidth]{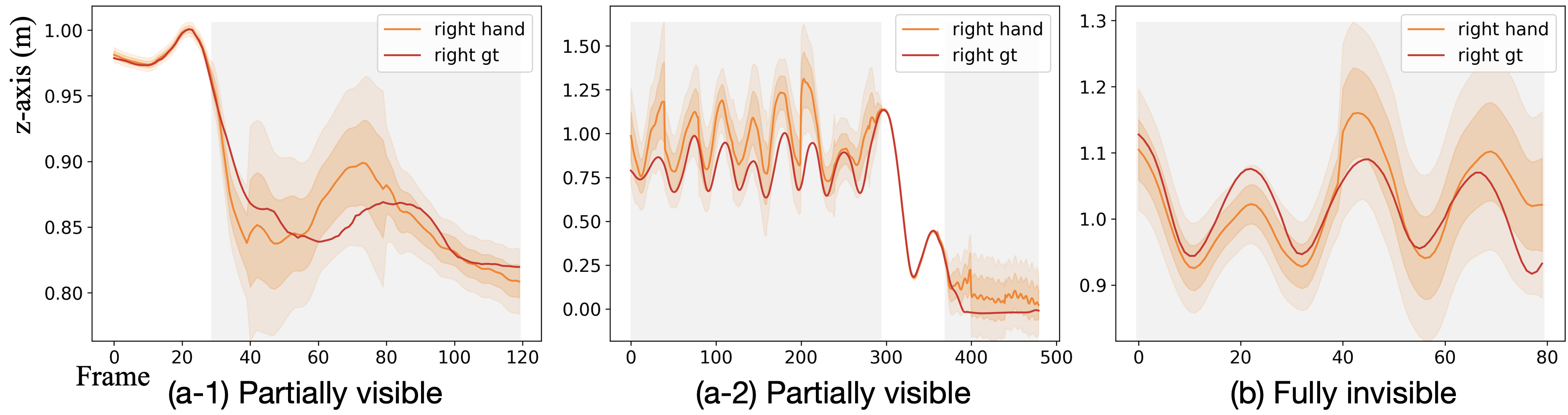}
    \caption{Uncertainty visualization of the right hand pose captured by the MAE. Gray areas represent frames where the hand is invisible, and white areas denote visible frames. We depict aleatoric uncertainty within ranges of $\pm1\sigma$ and $\pm2\sigma$ from the estimated $\mu$.}
    \label{fig:uncertainty}
\end{figure}

We visualized the aleatoric uncertainty in \Cref{fig:uncertainty}, captured by a model trained using MAE on the AMASS dataset. In cases of partial visibility, as shown in \Cref{fig:uncertainty} (a-1) and (a-2), the uncertainty range is notably small. Conversely, in frames where the subject is completely obscured, the uncertainty range increases significantly. Even in fully invisible scenarios, the model captures a range of uncertainty, likely influenced by head movements. Most of the estimated frames fall within the $\pm 2\sigma$ range.

\begin{figure}[t]
    \centering
    \begin{minipage}{0.5\textwidth}
        \subfloat[Video Frames]
        {\includegraphics[width=\linewidth, height=1.5cm]{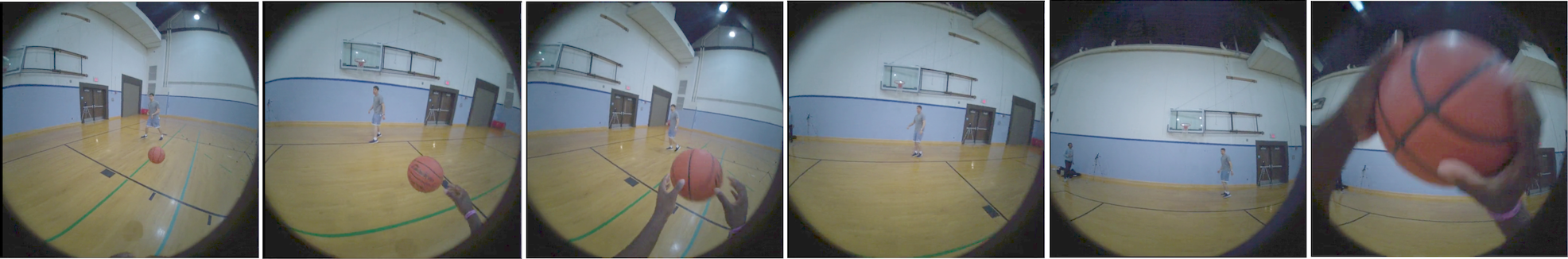}} \\
        \subfloat[Ego-Exo4D Skeleton Groundtruth and Prediction]
        {\includegraphics[width=\linewidth, height=2.5cm]{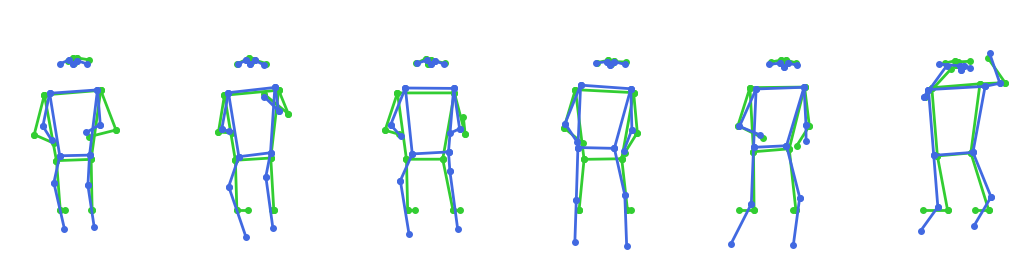}}
    \end{minipage}
    \hfill
    \begin{minipage}{0.45\textwidth}
        \centering
        \subfloat[AMASS Groundtruth and Prediction]
        {\includegraphics[width=\linewidth, height=5.0cm]{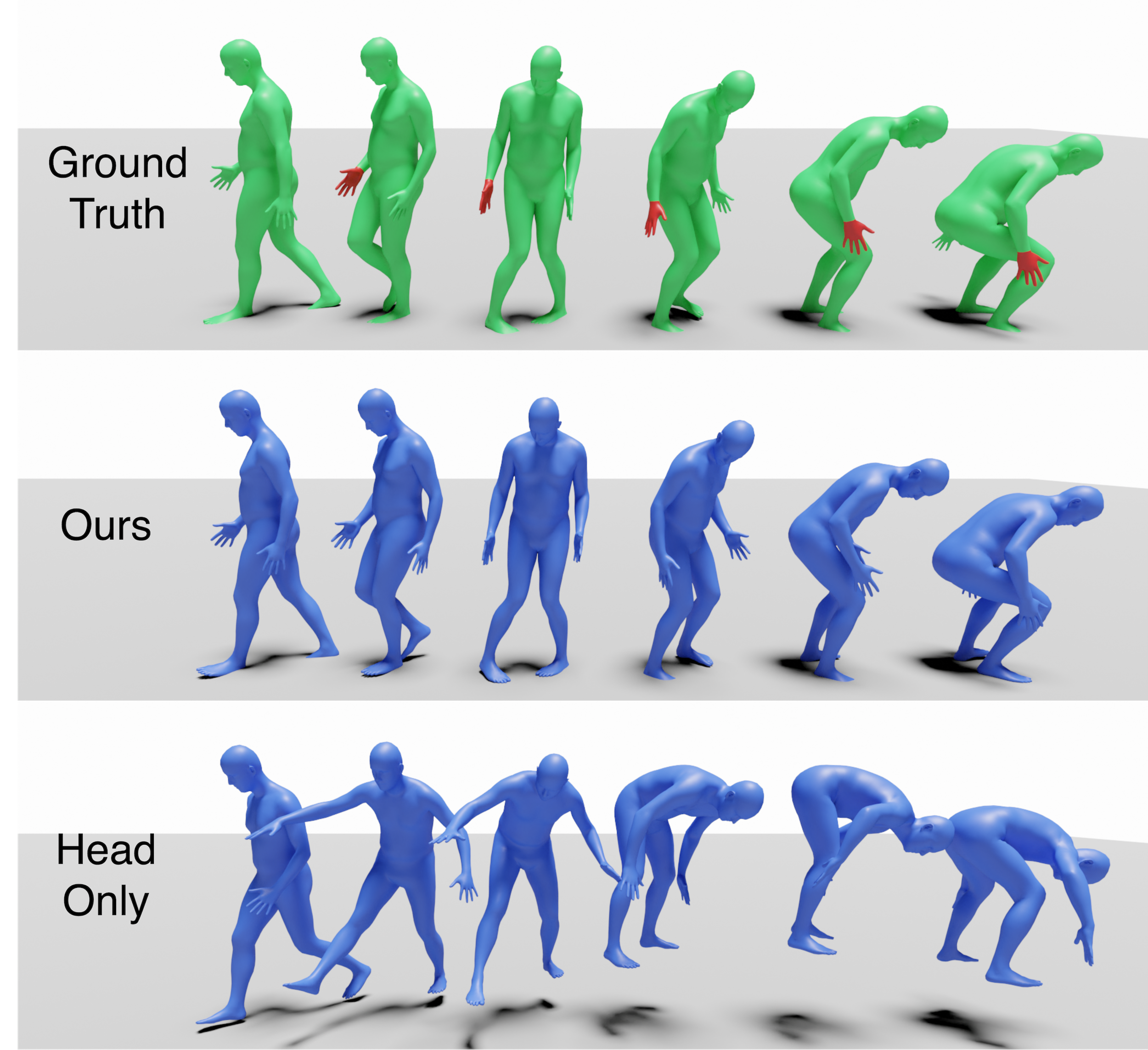}}
    \end{minipage}
     \caption{(a) Ego-Exo4D video frames, (b) the corresponding skeleton ground truth and our prediction results, and (c) qualitative results on AMASS data under different input conditions. \gr{green} indicates the ground truth, \bl{blue} indicates the predicted result, and \rd{red} indicates the visible hands. Head only estimates body pose from head trajectories, whereas Ours estimates body pose from imputed hand and head trajectories.} 
    \label{fig:qual}
\end{figure}

We also visualized the qualitative results on the Ego-Exo4D dataset and AMASS dataset in \Cref{fig:qual}. The qualitative results for AMASS show that our method improves the estimation results when sparse hand information is available, compared to the Head Only results. Additionally, in the Ego-Exo4D results, the hands are more aligned compared to the lower body when hands are available.

\section{Related Works}
\paragraph{Human Pose Estimation from Sparse Input} A common capture setting in mixed reality involves using a head-mounted device and hand controllers. Estimating full-body motion from the sparse input of head and hand movements is challenging. Recently, several methods have been proposed to tackle this:
AvatarPoser \cite{jiang2022avatarposer} is the first learning-based method to predict full-body poses in world coordinates using only head and hand motion inputs. It uses a Transformer encoder to extract deep features and decouples global motion from local joint orientations, refining arm positions with inverse kinematics for accurate full-body motion.
BoDiffusion \cite{castillo2023bodiffusion} employs a generative diffusion model for motion synthesis, addressing the under-constrained reconstruction problem. It uses a time and space conditioning scheme to leverage sparse tracking inputs, generating smooth and realistic full-body motion sequences.
AvatarJLM \cite{zheng2023realistic} uses a two-stage framework where sparse signals are embedded into high-dimensional features and processed by an MLP to generate joint-level features. These features are then converted into tokens and fed into a transformer-based network to capture spatial and temporal dependencies, with an SMPL regressor transforming them into 3D full-body pose sequences.
HMD-poser \cite{dai2024hmd} combines a lightweight temporal-spatial feature learning network with regression layers and uses forward kinematics to achieve real-time human motion tracking.
AGRoL~\cite{du2023avatars} utilized conditional diffusion model to generate full body pose from sparse upper-body tracking signals.
It is worth noting a concurrent work, EgoPoser~\cite{jiang2023egoposer}, which also addresses ego body pose estimation from doubly sparse observations. Their focus lies in preparing training data through field-of-view (FoV) modeling rather than introducing new algorithms. Our work is orthogonal to theirs, providing algorithmic contributions through a multi-stage pipeline including an uncertainty-aware masked auto-encoder (MAE).

\paragraph{Human Body Pose Estimation from Egocentric Videos} Estimating full 3D human body pose from egocentric videos is an ill-posed problem due to the partial visibility of wearer's body parts from the camera mounted on wearer's head. Recently, several approaches have been proposed to address this challenge. EgoEgo \cite{li2023ego} integrates SLAM and a learned transformer to estimate head motion, then leverages estimated head pose to generate plausible full-body motions using diffusion models. \cite{luo2021dynamics} designs a kinematic policy to generate per-frame target motion from egocentric inputs, and leverages a pre-learned dynamics model to distill human dynamics information into the kinematic model. GIMO \cite{zheng2022gimo} integrates motion, 3D eye gaze, and 3D scene features to generate gaze informed long term intention-aware human motion prediction. \cite{wang2022estimating} leverages external camera to generate pseudo labels to estimate full 3D body pose from single head mounted fish eye camera using weak supervision. \cite{wang2023scene} estimates geometry of surrounding objects and extracts 2D body pose features using EgoPW \cite{wang2022estimating} to regress 3D body pose with a voxel-to-voxel network \cite{moon2018v2v}.
\section{Conclusion}
\label{sec:conclusion}
In this paper, we have addressed the problem of egocentric body pose estimation using temporally sparse observations from head-mounted displays (HMDs). By leveraging both temporal and spatial completion, our approach effectively utilizes intermittent hand pose detections from egocentric videos, alongside consistently available head pose data, to reconstruct full-body motions. Through comprehensive experiments on datasets such as AMASS and Ego-Exo4D, we have demonstrated the effectiveness of our framework. Our results indicate significant improvements over existing methods, particularly in scenarios where dense sensor data may not be available or practical. This advancement opens up new possibilities for beneficial augmented reality experiences in various applications, including sports training by providing feedback on body mechanics, and other scenarios where users need to move freely without additional sensors such as hand controllers. However, our method has not been explicitly tested for fairness across different demographic groups. Potential biases in the datasets used could result in uneven performance across various user populations. Careful curation of training datasets is necessary to prevent unfair failures for underrepresented groups.

\section{Limitations}
\label{sec:limitations}
While our proposed method for estimating the body movements of a camera wearer from sparse tracking signals shows promising results, several limitations should be acknowledged. Firstly, our method has been tested with only one type of sparse body part tracking signal, specifically the hand. Incorporating the detection of other body parts, such as feet and elbows, may improve overall body pose estimation. Additionally, variations in lighting, occlusions, and the quality of the egocentric video can impact the accuracy of hand pose detection, subsequently affecting the overall body pose estimation.

The effectiveness of our method was validated using the AMASS and Ego-Exo4D datasets. Although these datasets are comprehensive, they may not encompass the full spectrum of possible real-world variations. Our study focused on pose estimation within a window size of less than a few seconds, following standard settings from the literature. It remains unclear how our method will perform with larger window sizes. Furthermore, the scalability of our method with larger datasets has not been thoroughly evaluated. The use of diffusion models for pose estimation may limit its utility for real-time applications due to their inference speed. Additionally, using multiple models to compute epistemic uncertainty can be computationally intensive.

\paragraph{Acknowledgement}
We acknowledge Feddersen Chair Funds and the US National Science Foundation (FW-HTF 1839971, PFI-TT 2329804) for Professor Karthik Ramani. Any opinions, findings, and conclusions expressed in this material are those of the authors and do not necessarily reflect the views of the funding agency. We sincerely thank the reviewers for their constructive suggestions.

\label{sec:acknowledgement}

\bibliography{main}


\appendix
\newpage

\section{Additional Details}

\subsection{Architecture \& Experimental Details} \label{sec:details}

\begin{table}[h]
    \centering
    \begin{minipage}{0.53\textwidth}
        \centering
        \resizebox{\textwidth}{!}{
            \begin{tabular}{c|ccc}
                \hline
                \textbf{Module} & \textbf{\# of Params} & \textbf{MACs} & \textbf{Time} \\
                \hline
                VQ-VAE          & 17.9 M                & 3.6 G         & 3 ms          \\
                MAE             & 51.3 M                & 23.3 G        & 4 ms          \\
                VQ-Diffusion    & 74.2 M                & 1190.2 G      & 958 ms        \\
                \hline
            \end{tabular}
        }
        \vspace{0.2em}
        \caption{Experimental results for different modules}
        \label{tab:module_results}
    \end{minipage}
    \hfill
    \begin{minipage}{0.46\textwidth}
        \centering
        \resizebox{\textwidth}{!}{
            \begin{tabular}{c|c|cccc|c}
                \multicolumn{2}{c|}{} & \multicolumn{4}{c|}{\textbf{Training step}} & \textbf{Times}\\
                \cline{3-6}
                \multicolumn{2}{c|}{} & 25 & 33 & 50 & 100 & (ms)\\
                \hline
                \multirow{4}{*}{\makecell{\rotatebox{90}{\textbf{Inference}}}} & 25  & 5.83 & 5.92 & 5.69 & 8.72 & 278\\
                & 33  & -    & 5.67 & 5.63 & 5.58 & 348 \\
                & 50  & -    & -    & 5.61 & 5.53 & 522 \\
                & 100 & -    & -    & -    & 5.51 & 1013 \\
                \hline
            \end{tabular}
        }
        \caption{Performance results for different inference and training step combinations}
        \label{tab:infer_train_steps}
    \end{minipage}
\end{table}
\paragraph{VQ-VAE}
We adhered to the architectural details and training protocol of Zhang et al. \cite{zhang2023generating}, with modifications including setting both the encoder and decoder stride to 1, and adjusting the window size to 40. For the Ego-Exo4D dataset, we employed wing loss with a width of 5 and a curvature of 4. For AMASS, we opted for L2 loss. 
Additionally, to generate smooth motion, we applied both velocity and acceleration losses, assigning weights of 10 for each in the AMASS dataset, and weights of 1 for each in the Ego-Exo4D dataset.
The input shape is represented as $T_w \times J \times D_{\text{data}}$, where $J$ is the number of joints, $T_w$ is time window, and $D_{\text{data}}$ is the data representation. 
For AMASS, which uses a 6D rotation representation, $J = 22$ and $D_{data} = 6$. For Ego-Exo4D, $J = 17$ and $D_{data} = 3$. We use $T_w = 40$ for both AMASS and Ego-Exo4D.

\paragraph{VQ-Diffusion}
We employed the same hyperparameters and training specifications as Gu et al. \cite{gu2022vector} for training VQ-Diffusion. 
Additionally, we replaced the absolute positional encoding with relative positional encoding, following the implementation of VQ-Diffusion for human motion generation proposed by Chi et al. \cite{chi2024m2d2m}.
We replaced the text condition module with the uncertainty-aware MAE module to feed the imputed trajectory and uncertainty as a conditional input.

\paragraph{Masked Auto-Encoder}
We adapted the encoder to accommodate a variable number of visible hands and modified the last projection layer to guide the uncertainty, as detailed in the main paper. Otherwise, we followed the training details provided by He et al. \cite{he2022masked}. We trained 4 models to measure the uncertainty, $M = 4$.

\paragraph{Dataloader}
We adopted the dataloader of AMASS and evaluation configurations from Bodiffusion \cite{castillo2023bodiffusion} for our experiments. For the Ego-Exo4D dataset, we employed the dataset implementation from the Ego-Exo4D \cite{grauman2023ego}.

\paragraph{Analaysis on Computaional Cost}
We evaluated the computational cost of our approach by analyzing the number of parameters, multiply-accumulate operations (MACs), and inference time for each module in our pipeline. The reported times in \Cref{tab:infer_train_steps} represent the total measured time for the entire pipeline, whereas \Cref{tab:module_results} measures the time only for the corresponding module, excluding overhead between modules. As shown in \Cref{tab:module_results}, the VQ-Diffusion model is responsible for the majority of MACs and inference time. To mitigate this, we conducted further experiments to explore ways of reducing the VQ-Diffusion model's inference time. \Cref{tab:infer_train_steps} presents the trade-off between performance and inference time based on the number of diffusion steps, offering multiple options. Notably, training with 50 steps and inferring with 25 steps yields approximately a 4x faster inference time with only about 3\% reduction in performance.

\section{Compute Resource} 
\label{sec:compute}
We ran our experiments on one workstation, containing AMD Ryzen Threadripper PRO 7975WX, DDR5 RAM 256GB and 4 NVIDIA GeForce RTX 4090. AMASS dataset takes 512GB of storage and Ego-Exo4D dataset takes 11TB. One training run took around 18 hours on 1 GPU for AMASS and around 12 hours for Ego-Exo4D. Inference over AMASS validation set takes 40 minutes on 1 GPU and inference over Ego-Exo4D validation set takes 10 minutes on 1 GPU. In total, all experiments including preliminary or failed experiments took approximately 300 GPU-hours.

\section{Licenses for Assets Used in the Paper} \label{sec:license}
\paragraph{Code} We use the code of BoDiffusion \cite{castillo2023bodiffusion} which is available at \href{https://github.com/BCV-Uniandes/BoDiffusion}{https://github.com/BCV-Uniandes/BoDiffusion}. Unfortunately, we could not locate the licensing terms for the source code. For the Masked Auto Encoder, we use the implementation available at \href{https://github.com/pengzhiliang/MAE-pytorch}{https://github.com/pengzhiliang/MAE-pytorch}, but we could not find the licensing terms for this source code.

We also employ VQ-Diffusion \cite{gu2022vector}, available at \href{https://github.com/cientgu/VQ-Diffusion/tree/main?tab=readme-ov-file}{https://github.com/cientgu/VQ-Diffusion/tree/main?tab=readme-ov-file}, which is licensed under Microsoft's Open Source Program.

For VQ-VAE, we use the implementation from T2M-GPT \cite{zhang2023generating}, which can be found at \href{https://github.com/Mael-zys/T2M-GPT}{https://github.com/Mael-zys/T2M-GPT}, and is licensed under the Apache License 2.0.

For 3D hand detection, we use the code of FrankMoCap \cite{rong2021frankmocap}: \href{https://github.com/facebookresearch/frankmocap}{https://github.com/facebookresearch/frankmocap}, which is licensed under the CC BY-NC 4.0 license. We also used RTM-pose \cite{jiang2023rtmpose}, which is available at \href{https://github.com/open-mmlab/mmpose}{https://github.com/open-mmlab/mmpose} under the Apache License 2.0.

\paragraph{Data} We use the Ego-Exo4D dataset \cite{grauman2023ego} \href{https://ego-exo4d-data.org}{https://ego-exo4d-data.org}, which is licensed under a \href{https://ego4d-data.org/pdfs/Ego-Exo4D-Model-License.pdf}{custom (commercial or non-commercial) license}.
We also use AMASS \cite{mahmood2019amass} \href{https://amass.is.tue.mpg.de}{https://amass.is.tue.mpg.de}, which is licensed under a \href{https://amass.is.tue.mpg.de/license.html}{custom (non-commercial scientific research) license}.

\section{Preliminary}
\label{sec:prelim}
\subsection{Discrete Diffusion Model}
Discrete diffusion models \cite{gu2022vector} represent a category of diffusion models that progressively introduce noise into data while training to reverse this process. In contrast to continuous models, such as a latent diffusion model \cite{rombach2022high}, which manipulate data in a continuous state space, discrete diffusion models operate within discrete state spaces.

\paragraph{VQ-VAE}\label{sec:vqvae}
Vector Quantized-Variational Autoencoder (VQ-VAE) \cite{van2017neural} is a generative model that extends the concept of Variational Autoencoders (VAEs) by incorporating discrete latent representations via vector quantization. The encoder $ E(\mathbf{x}) $ compresses input data $ \mathbf{x} $ into discrete latent vectors by mapping each encoded representation to the closest vector $ \mathbf{z}_q $ to the nearest codebook entry from a learned codebook of prototypes using the nearest-neighbor search: $\mathbf{z}_q\!=\!Q(\mathbf{z})\!=\!\text{argmin}_{\textbf{c}_i \in \mathcal{C}}{||\mathbf{z} - \textbf{c}_i||_2}$. Here, $\mathcal{C}\!=\!\{\textbf{c}_1, \ldots, \textbf{c}_K\}$, where $ K $ is the total number of codebooks. The decoder $ D(\mathbf{z}_q) $ reconstructs the input data $ \mathbf{x}$ from these quantized vectors, yielding a reconstructed output $\hat{\mathbf{x}} = D(\mathbf{z}_q)$.
The optimization process involves minimizing a combination of reconstruction loss and commitment loss. The reconstruction loss is expressed as $ \| \mathbf{x} - \hat{\mathbf{x}} \|_2 $, while the commitment loss ensures the encoder commits to the nearest prototype in the codebook: $ \| \text{sg}[ \mathbf{z_q}] - \mathbf{z} \|_2 $, where $ \text{sg} $ is the stop-gradient operator. 
The overall loss function, which the VQ-VAE model minimizes, is: $ \mathcal{L}_{\text{VQ}} = \| \mathbf{x} - \hat{\mathbf{x}} \|_2 + \| \mathbf{z_q} - \text{sg}[ \mathbf{z} ] \|_2 + \lambda_{\text{VQ}} \| \text{sg}[ \mathbf{z_q} ] - \mathbf{z} \|_2 $.
Here, $\lambda_{\text{VQ}}$ is a coefficient for the commitment loss.

\paragraph{Forward Diffusion Process.}
Building on the foundation laid by the discrete diffusion models introduced by \cite{sohl2015deep}, VQ-Diffusion \cite{gu2022vector} refined the diffusion process with a mask-and-replace strategy. In VQ-Diffusion, during the forward diffusion process, tokens can either transition to other tokens or to a special \texttt{<MASK>} token.
The transition probability from token $\mathbf{z}^i$ to $\mathbf{z}^j$ at diffusion step $t$ is defined by the matrix $\mathbf{Q}_t[i,j]$. The transition matrix $\mathbf{Q}_t$, structured in $\mathbb{R}^{(K+1) \times (K+1)}$, follows:
\begin{flalign}
\mathbf{Q}_t =
\setlength\arraycolsep{0.7em}
\mleft[
\begin{array}{c|c}
  \hat{\mathbf{Q}}_t & 0 \\
  \hline
  \gamma_t\cdot\mathbf{1}^\top & 1
\end{array}
\mright], \text{where } \hat{\mathbf{Q}}_t = \alpha_t\mathbf{I}+\beta_t\mathbf{1}\mathbf{1}^\top.
\label{eq:q_t}
\end{flalign}
Here, $\alpha_t$ adjusts to ensure conservation of probability, such that $\alpha_t\!=\!1\!-\!K\beta_t\!-\!\gamma_t$ is the probability of transitioning between tokens, and $\gamma_t$ governs transitions to the \texttt{<MASK>} token.
The transition from step $t-1$ to $t$ is expressed as:
$q(\mathbf{z}_{t}|\mathbf{z}_{t-1}) = \boldsymbol{v}^{\top}(\mathbf{z}_t)\boldsymbol{Q}_t\boldsymbol{v}(\mathbf{z}_{t-1}),$
where $ \boldsymbol{v}(\mathbf{z}_t) \in \mathbb{R}^{(K+1) \times 1}$ is an one-hot encoded vector representing the token index of $\mathbf{z}_t$.
Using the Markov property, the probability of transitioning from any initial step $0$ to step $t$ is
$q(\mathbf{z}_{t}|\mathbf{z}_{0}) = \boldsymbol{v}^{\top}(\mathbf{z}_t)\overline{\boldsymbol{Q}}_t\boldsymbol{v}(\mathbf{z}_{0})$, 
where $\overline{\mathbf{Q}}_t = \mathbf{Q}_t \mathbf{Q}_{t-1}\cdots \mathbf{Q}_1$ is the cumulative transition matrix. This defines the cumulative probabilities as $\bar{\alpha}_t = \prod_{i=1}^{t} \alpha_i, \bar{\gamma}_t = 1 - \prod_{i=1}^{t} (1 - \gamma_i), \text{and } \bar{\beta}_t = (1 - \alpha_t - \gamma_t)/K$.

\paragraph{Conditional Denoising Process.} 
In the conditional denoising process, a neural network denoted as $p_{\theta}$ aims to predict the original, noiseless token $\mathbf{z}_0$ given a corrupted token and the associated condition, such as a embedded hand trajectories. The posterior distribution for the discrete diffusion process can be defined as:
\begin{flalign}
q(\mathbf{z}_{t-1}|\mathbf{z}_t,\mathbf{z}_0) &= \frac{q(\mathbf{z}_{t-1}|\mathbf{z}_0)q(\mathbf{z}_t|\mathbf{z}_{t-1},\mathbf{z_0})}{q(\mathbf{z}_t|\mathbf{z}_0)} \nonumber \\
&= \frac{\big(\boldsymbol{v}^{\top}(\mathbf{z}_t)\boldsymbol{Q}_t\boldsymbol{v}(\mathbf{z}_{t-1})\big)\big(\boldsymbol{v}^{\top}(\mathbf{z}_{t-1})\boldsymbol{\overline{Q}}_{t-1}\boldsymbol{v}(\mathbf{z}_{0})\big)}{\boldsymbol{v}^{\top}(\mathbf{z}_t)\boldsymbol{\overline{Q}}_t\boldsymbol{v}(\mathbf{z}_{0})}.\end{flalign}
With this, the reverse transition distribution is determined as:
\begin{flalign}
p_{\theta}(\mathbf{z}_{t-1} | \mathbf{z}_t, \mathbf{y}) = {\textstyle\sum}^{K}_{\tilde{\mathbf{z}}_0=1} q(\mathbf{z}_{t-1} | \mathbf{z}_t, \tilde{\mathbf{z}}_0) p_{\theta}(\tilde{\mathbf{z}}_0 | \mathbf{z}_t, \mathbf{y}),
\end{flalign}
where the network iteratively denoises tokens from step $T$  down to 1, eventually generating the token $ \mathbf{z}_0 $ conditioned on $ \mathbf{y} $.
To train the network $ p_{\theta} $, the training approach includes not only a denoising objective but also the standard variational lower bound objective \cite{sohl2015deep}, denoted as $ \mathcal{L}_{\text{vlb}} $. The comprehensive training objective is:
\begin{flalign}
\mathcal{L} = \mathcal{L}_{\text{vlb}} + \lambda \mathbb{E}_{\mathbf{z}_t \sim q(\mathbf{z}_t|\mathbf{z}_0)}[-\log{p_{\theta}(\mathbf{z}_0|\mathbf{z}_t,\mathbf{y})}],
\label{eq:vq_diffusion}
\end{flalign}
where $ \lambda $ is the coefficient for the denoising loss.

\section{Additional Experimental results}

\begin{figure}[h!]
    \centering
    \includegraphics[width=1.0\textwidth]{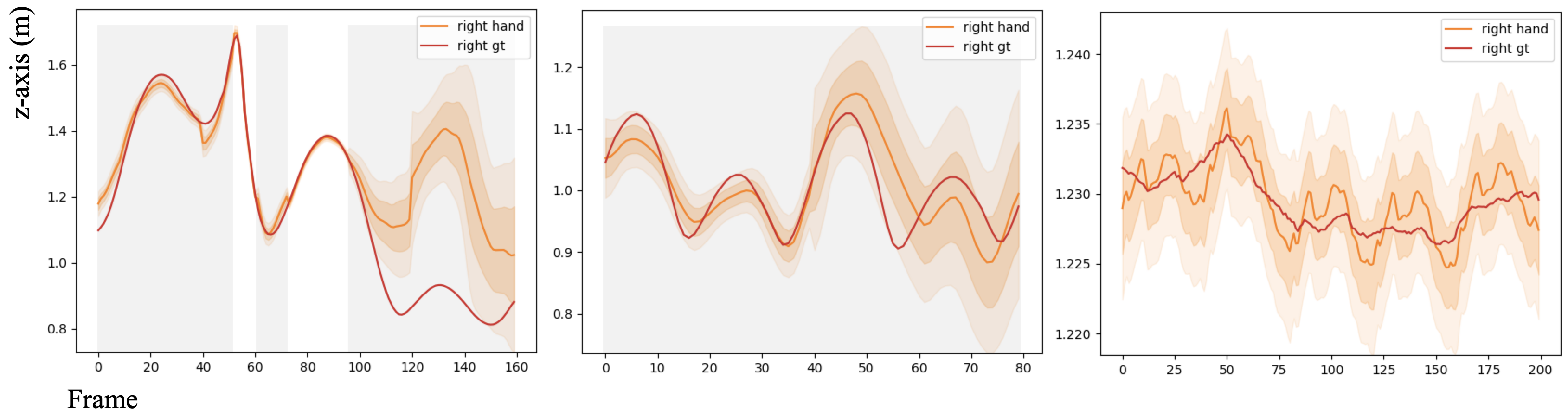}
    \caption{Additional uncertainty visualization of the right hand pose captured by the MAE. Gray areas represent frames where the hand is invisible, and white areas denote visible frames. We depict aleatoric uncertainty within ranges of $\pm1\sigma$ and $\pm2\sigma$ from the estimated $\mu$.}
    \label{fig:uncertainty2}
\end{figure}

\begin{figure}[htp!]
    \centering
    \includegraphics[width=\textwidth]{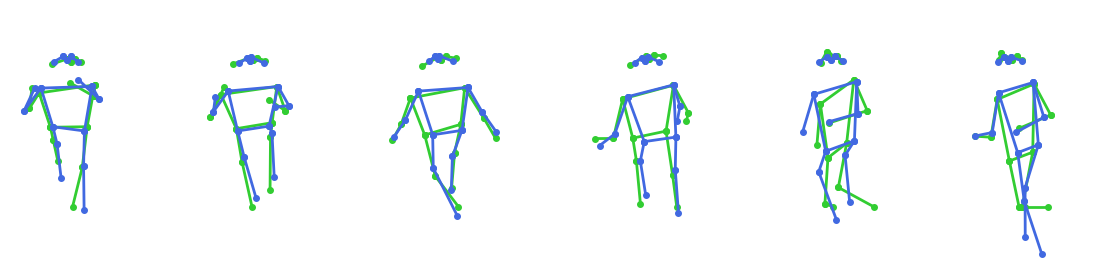}
    \label{fig:ego_exo_7}

    \vspace{-5pt}
    \includegraphics[width=\textwidth]{figs/37_1180_1240.png}
    \label{fig:ego_exo_8}

    \vspace{-5pt}
    \includegraphics[width=\textwidth]{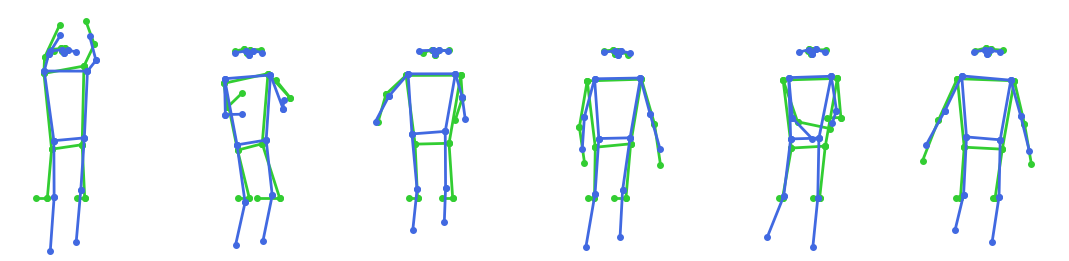}
    \label{fig:ego_exo_9}

    \vspace{-5pt}
    \includegraphics[width=\textwidth]{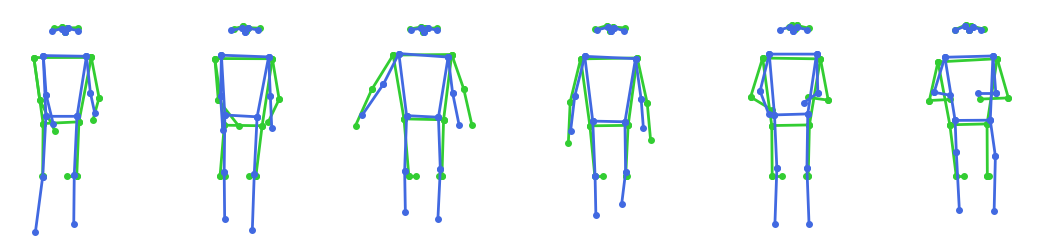}
    \label{fig:ego_exo_10}

    \vspace{-5pt}
    \includegraphics[width=\textwidth]{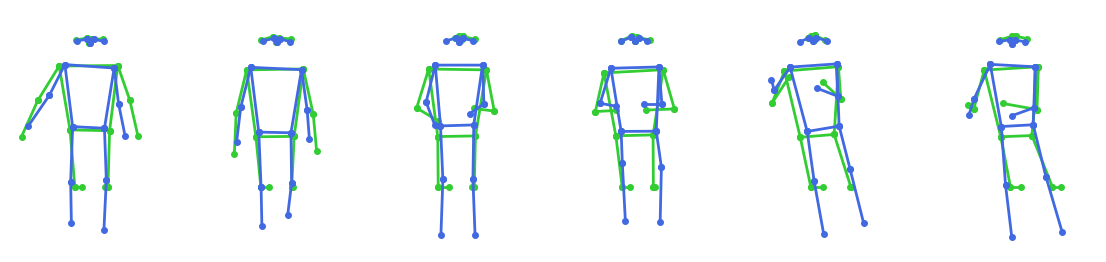}
    \label{fig:ego_exo_11}

    \vspace{-5pt}
    \includegraphics[width=\textwidth]{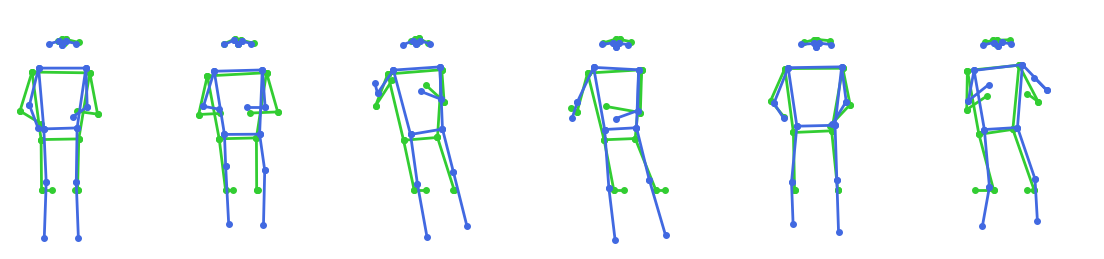}
    \label{fig:ego_exo_12}

    \caption{Qualitative results showing the groundtruth in \textcolor{green}{green} and predicted human pose in \textcolor{blue}{blue} using our method on the Ego-Exo4D dataset.}
    \label{fig:egoexo_skeletons}
\end{figure}

\begin{figure}[h!]
    \vspace{-0.5em}
    \centering
    \includegraphics[width=1.00\textwidth]{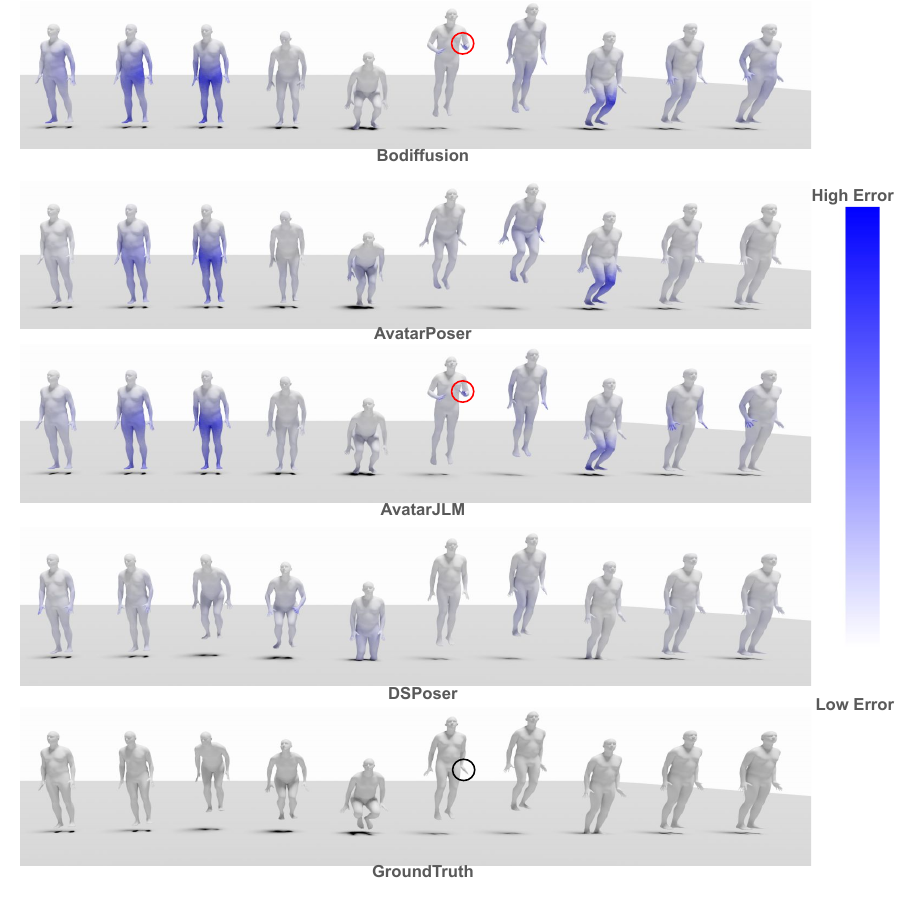}
        \caption{Qualitative results on AMASS dataset comparing DSPoser (Ours) against the baselines. Color gradient indicates an absolute positional error, with a higher error corresponding to higher blue intensity. Results demonstrate that motions generated by DSPoser exhibit greater similarity to the ground truth. Furthermore, it highlights higher errors (indicated with red circles) for baselines when the hand is occluded in the ground truth pose (indicated with a black circle). }
    \label{fig:qualitative_appendix}
    \vspace{-1.0em}
\end{figure}

\begin{figure}[h!]
    \centering
    \includegraphics[width=0.45\textwidth]{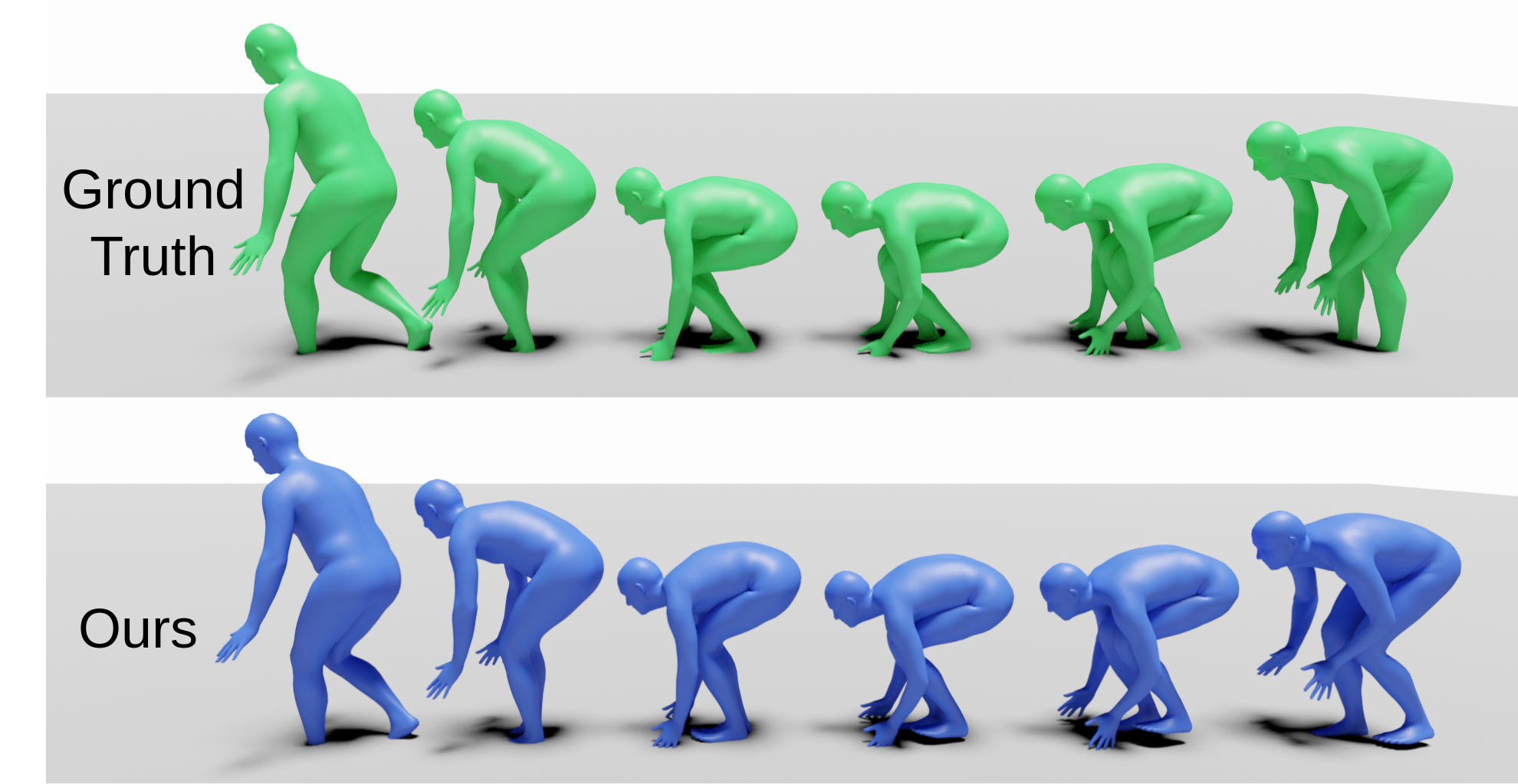}
    \hfill
    \includegraphics[width=0.45\textwidth]{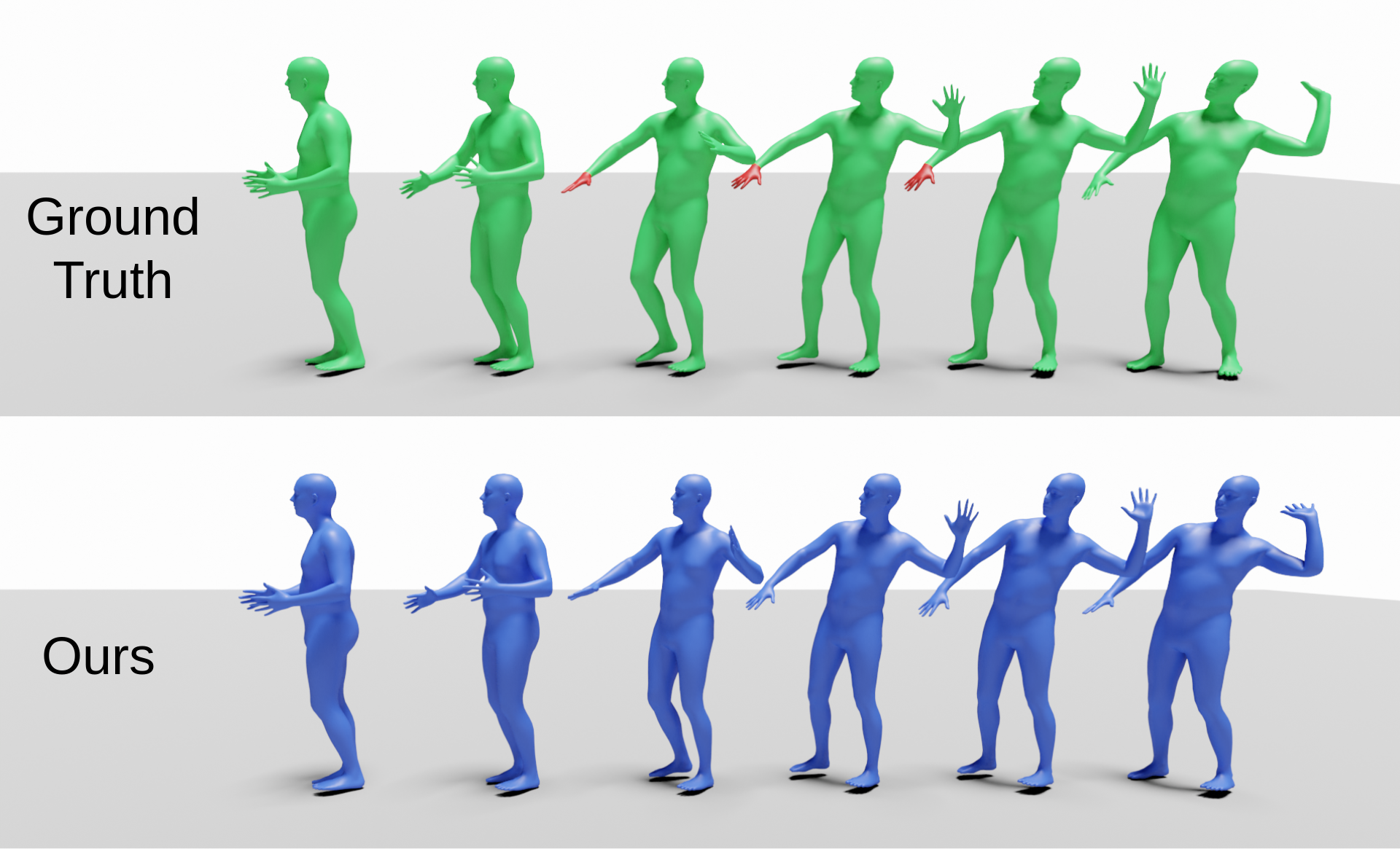}
    \hfill
    \includegraphics[width=0.45\textwidth]{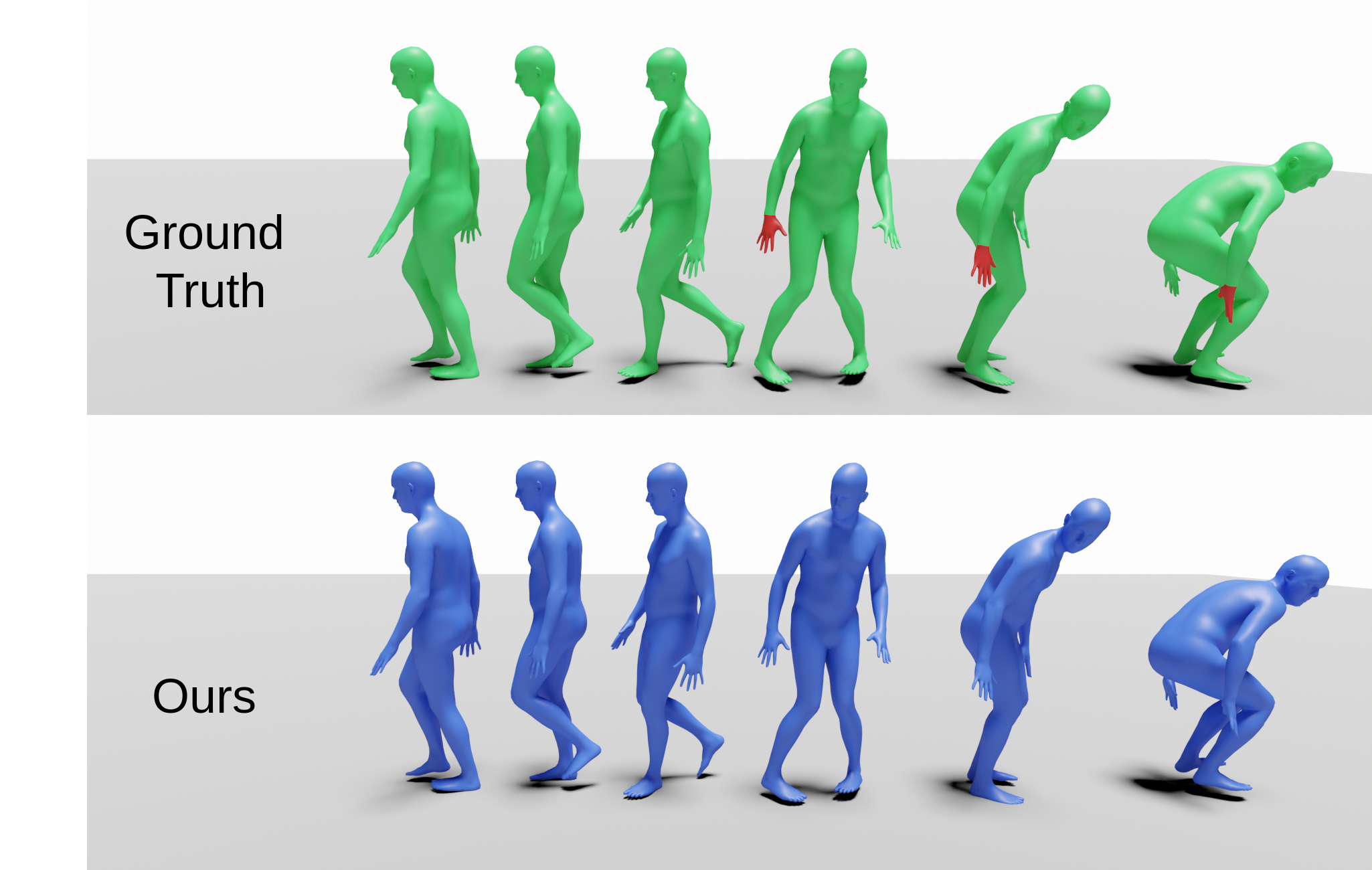}
    \hfill
    \includegraphics[width=0.45\textwidth]{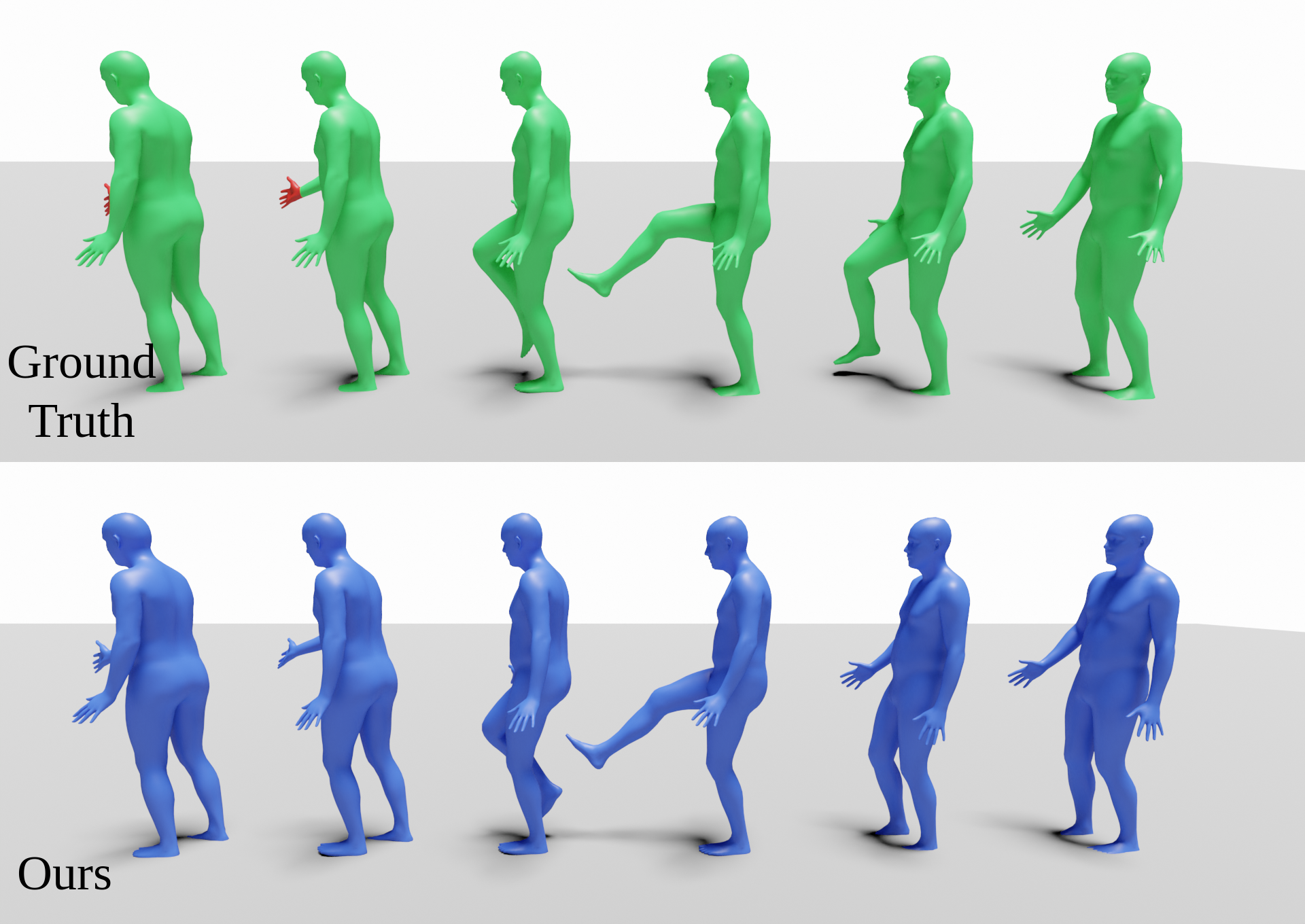}
    \caption{Qualitative results showing the groundtruth in \textcolor{green}{Green} and predicted human pose in \textcolor{blue}{blue} using our method on AMASS dataset, with \textcolor{red}{red} indicating the visible hands.}
    \label{fig:AMASS_meshes}
\end{figure}

We illustrate additional uncertainty visualization on Fig. \ref{fig:uncertainty2}.
In addition, we demonstrate the additional qualitative results with our method on Ego-Exo4D dataset and AMASS dataset in Fig. \ref{fig:egoexo_skeletons} and Fig. \ref{fig:qualitative_appendix} \& \ref{fig:AMASS_meshes}, respectively.

\newpage

\newpage
\clearpage


\end{document}